\documentclass[journal]{IEEEtran}
\usepackage{amsmath,amssymb,amsthm}
\usepackage{algorithm}
\usepackage{algpseudocode}
\usepackage{array}
\usepackage{booktabs}
\usepackage{multirow}
\usepackage{graphicx}
\usepackage{subcaption}
\usepackage{stfloats}
\usepackage{url}
\usepackage{textcomp}
\usepackage{balance}
\usepackage{hyperref}
\usepackage{xcolor}
\usepackage{tikz}
\usetikzlibrary{positioning,arrows.meta,fit,backgrounds,calc}
\hypersetup{colorlinks=true,linkcolor=blue!50!black,citecolor=blue!50!black,
    urlcolor=blue!60!black}

\newtheorem{proposition}{Proposition}
\theoremstyle{definition}
\newtheorem{definition}{Definition}

\newcommand{\relu}{\operatorname{ReLU}}
\DeclareMathOperator*{\argmin}{arg\,min}
\newcommand{\softmax}{\operatorname{softmax}}
\newcommand{\Kset}{\mathcal{K}}

\begin{document}
\title{Cascading Gradient Inversion via LT-Code Inspired Peeling in Federated Learning}

\author{Saeed~Shariati and Mohsen~Alambardar~Meybodi%
	\thanks{Both authors are with the Department of Applied Mathematics and Computer Science, 
		Faculty of Mathematics and Statistics, University of Isfahan, Isfahan 81746-73441, Iran 
		(e-mail: saeed.shariati1380@gmail.com; m.alambardar@sci.ui.ac.ir).}%
	\thanks{Corresponding author: Mohsen Alambardar Meybodi.}}


\maketitle

\begin{abstract}
	Federated learning shares model updates rather than raw data, yet these updates can be inverted to reconstruct the clients' training data. Analytic reconstruction attacks, which invert a gradient in closed form, degrade as the batch grows: prior single-round attacks recover only about half of a batch of size $100$ even when the attacker fully controls the network parameters, and known upper bounds limit what any such method can recover. We establish a connection between gradient inversion and the theory of erasure-correcting codes, and use it to construct attacks that exceed these bounds. Our attacks recover batches exactly, together with every sample's label, from a single FedSGD round, and certify each recovery without ground-truth data. On eight image and tabular benchmarks they outperform prior single-round attacks by a wide margin. Even a passive attacker who only observes an honestly trained network recovers $94$--$100\%$ of ImageNet batches at sizes up to $128$, more than prior single-round attacks achieve even with active manipulation of the model, and in the active setting more than $90\%$ is recovered at batch sizes of several hundred. These results show that the privacy leakage of federated learning has been underestimated.
\end{abstract}


\section{Introduction}\label{sec:introduction}

Federated learning (FL) enables collaborative training of machine learning models without requiring participants to share their raw data~\cite{mcmahan2017communication,kairouz2021advances}. In each communication round, a central server distributes the current global model to a subset of clients; each client computes an update on its private local data and returns only the update, typically a gradient or model difference, to the server. Two widely studied algorithms are FedSGD, in which clients return the gradient of a local batch, and FedAvg, in which clients perform multiple local stochastic gradient steps before returning the net model update~\cite{mcmahan2017communication}. FL deployments are further divided into cross-silo and cross-device settings~\cite{kairouz2021advances}: cross-silo clients are organizations that remain available throughout training, whereas cross-device clients are numerous mobile or IoT devices, only a fraction of which participate in each round and any of which may appear only once.

Although FL was designed with privacy in mind, numerous studies have shown that the shared updates can leak substantial information about clients’ training data. Such leakage has given rise to a variety of privacy attacks, which are commonly grouped into three categories: data reconstruction~\cite{zhu2019deep,zhao2020idlg,Geiping2020Inverting,Boenisch2023Curious,diana2025cutting,Diana2026NoGuessing}, membership inference~\cite{Nasr2019Comprehensive,Melis2019Exploiting}, and property inference~\cite{Ganju2018Property,Melis2019Exploiting}. These attacks may be either \emph{active}, where the server deliberately modifies weights, biases, or even the model architecture to facilitate reconstruction~\cite{Fowl2022Robbing,Pasquini2022Eluding,Wen2022Fishing}, or \emph{passive}, where an honest-but-curious server attempts to recover information solely by inspecting the received updates~\cite{Dimitrov2024SPEAR,Bakarsky2025SPEARpp}. In response, several defenses have been proposed, most notably differential privacy (DP)~\cite{Dwork2006Differential} and secure aggregation (SA)~\cite{Bonawitz2017SecureAgg,Hosseini2025SecureAgg}. However, secure aggregation alone has been shown to be insufficient against certain privacy attacks, such as membership inference, and is therefore typically combined with differential privacy~\cite{Ngo2024SecureAgg}.

This paper focuses on \emph{data reconstruction} attacks. Within this family one can further distinguish optimization-based~\cite{zhu2019deep,zhao2020idlg,Geiping2020Inverting,Qian2024GISMN,Lu2022APRIL}, generative~\cite{jeon2021gradient,zhang2023generative}, and analytic~\cite{Boenisch2023Curious,diana2025cutting,Zhang2023Compromise,Diana2026NoGuessing,Shi2023ScaleMIA,Wen2022Fishing} methods. Analytic attacks exploit explicit mathematical relationships between model gradients and the underlying training data. A prominent class of such attacks exploits closed-form relationships involving the first-layer gradients and input samples, typically for networks whose first layer is fully connected and followed by a ReLU activation. When a neuron is activated by only a single sample in the batch, that sample can be recovered exactly by a simple ratio of the corresponding weight and bias gradients~\cite{Boenisch2023Curious}. Active analytic attacks are often easily detectable due to modifications of the model architecture~\cite{Fowl2022Robbing,Zhao2024Loki} or conspicuous changes in model parameters~\cite{diana2025cutting,Shi2023ScaleMIA,Wen2022Fishing}, while passive analytic attacks have so far struggled with large batch sizes~\cite{Dimitrov2024SPEAR}.

\begin{figure}[t]
    \centering
    \includegraphics[width=\columnwidth]{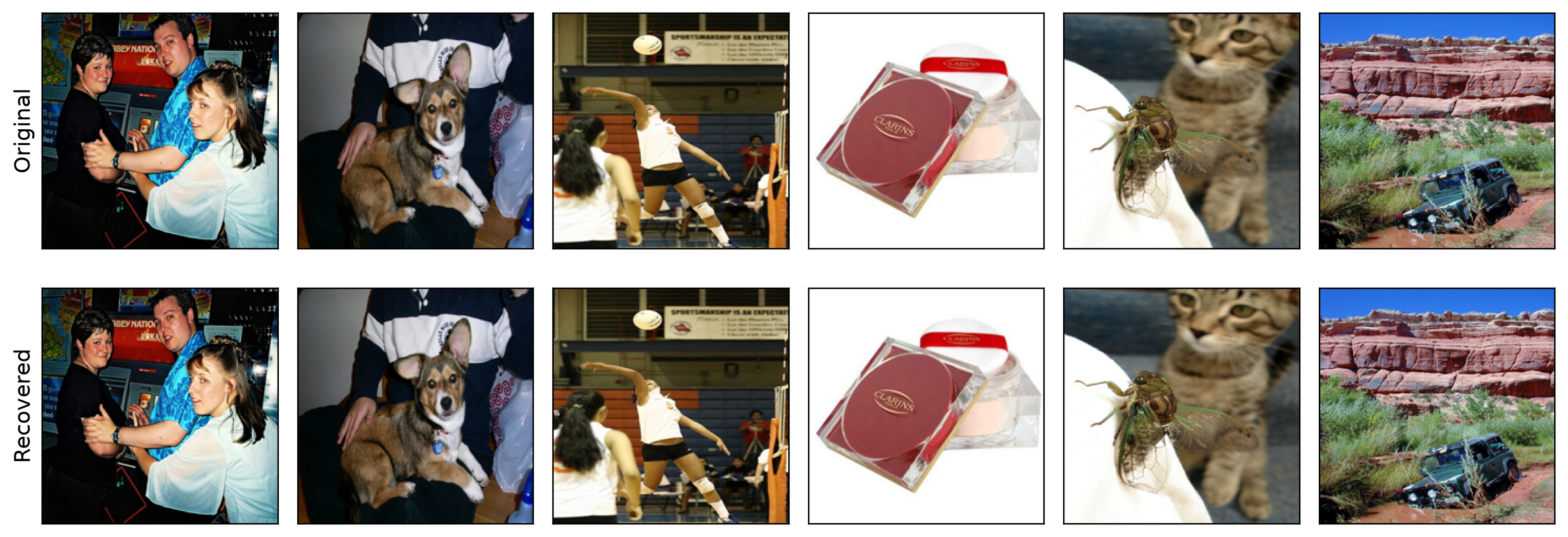}
    \caption{Original (top) and recovered (bottom) ImageNet samples from a single round of FedSGD.}
    \label{fig:teaser}
\end{figure}

Analytic attacks that rely on isolating individual samples, however, are fundamentally limited. Because they recover only isolated samples, the number of recoverable inputs is upper-bounded by the number of vertices of the convex hull formed by the batch~\cite{diana2025cutting}. This bound applies to isolation-based methods; it does not cover analytic attacks that recover batches through other mechanisms, such as the low-rank gradient structure exploited by SPEAR~\cite{Dimitrov2024SPEAR}. For data drawn from continuous distributions, the bound grows slowly with batch size and becomes particularly restrictive for low-dimensional (e.g., tabular) inputs. Consequently, single-round isolation-based methods recover only a fraction of larger batches even under active model manipulation: with trap weights and $1000$ neurons, Boenisch et al.~\cite{Boenisch2023Curious} recover roughly half of a batch of $100$ samples, and the fraction decreases as the batch grows. Multi-round variants~\cite{diana2025cutting,Diana2026NoGuessing} recover full batches, but require repeated interaction with the client under carefully controlled per-round coefficients; this rules them out in the cross-device setting, where a client may be present in only a single round.

In this work we show that the limitation is not intrinsic. Once an isolated sample is recovered, its contribution can be subtracted from the observed gradient. The residual gradient is again of the same analytic form, but now computed over the remaining unrecovered samples. This subtraction can create new isolated samples, which can themselves be recovered and subtracted, producing an iterative cascade. The process is formally analogous to the peeling decoder of Luby Transform (LT) codes~\cite{Luby2002LT} (Section~\ref{sec:bg-lt}).

To increase the length and success probability of the cascade, we further control the neuron degrees (Definition~\ref{def:degree}) of the first-layer neurons. Adapting the degree distribution of LT codes, we shape these degrees toward the robust soliton distribution. We introduce two novel constructions: Soliton-Free, which requires no auxiliary data and relies on a Gaussian approximation of pre-activations, and Soliton-Data, which uses a single auxiliary batch to set exact degree thresholds. Both constructions primarily adjust the first-layer biases to control neuron degrees.

Our attacks recover large batches exactly, together with their labels, from a single FedSGD communication round. The label of each recovered sample is obtained from the residual bias via a closed-form fit that also serves as a certificate of authenticity (Section~\ref{sec:certify}); no ground-truth data are required. The methods apply to any classification network whose first layer is a fully connected layer with ReLU activation, trained with cross-entropy loss (though they can be extended to a broader class of networks~\cite{Boenisch2023Curious,Zhang2023Compromise}), and they succeed on both image and tabular data.

\begin{itemize}
	\item We introduce a cascading recovery procedure for analytic gradient inversion that iteratively recovers and subtracts isolated samples, overcoming the upper bound that limits prior isolation-based attacks (Section~\ref{sec:related}).
	\item We propose two parameter-manipulation techniques, Soliton-Free and Soliton-Data, that shape neuron activation degrees according to the robust soliton distribution, substantially increasing the fraction of recoverable samples. Soliton-Data recovers more than $90\%$ of samples at batch sizes of several hundred on many data sets, and Soliton-Free recovers $98\%$ of samples at $B=300$ without auxiliary data (Table~\ref{tab:recall_avg_C5_S99}).
	\item In the passive (honest-but-curious) setting, the attack achieves near-complete recovery ($94$--$100\%$) at batch sizes up to $128$ on image benchmarks and recovers a complete tabular batch at $B=64$. For comparison, the passive state of the art SPEAR++~\cite{Bakarsky2025SPEARpp} reconstructs batches of at most about $65$ samples at the same network width ($N=1000$) and fails at $B=150$, while prior single-round attacks failed to recover a batch size of $128$ even in active settings~\cite{Boenisch2023Curious}.
	\item Recovered samples are exact and paired with their correct labels, with perfect label accuracy across eight data sets, and every recovery is certified without access to ground-truth data.\end{itemize}

\section{Background}
\label{sec:background}

We write vectors in bold lowercase ($\mathbf{x}$) and matrices in bold uppercase ($\mathbf{X}$). Samples are indexed by $i$ and neurons by $j$. Table~\ref{tab:notations} summarizes the notation used throughout the paper.

\begin{table}[H]
	\centering
	\caption{Notation}
	\label{tab:notations}
	\setlength{\tabcolsep}{4pt}
	\footnotesize
	\begin{tabular}{ll}
		\toprule
		Symbol & Meaning \\
		\midrule
		$F$                    & Input dimension \\
		$\mathbf{x} \in \mathbb{R}^{F}$ & A single input sample \\
		$y \in \{1,\dots,K\}$  & Label of $\mathbf{x}$; $K$ = number of classes \\
		$\mathbf{X} \in \mathbb{R}^{B \times F}$ & Batch of $B$ samples (one per row) \\
		$B$                    & Batch size \\
		$N$                    & Number of first-layer neurons \\
		$\mathbf{W}_1 \in \mathbb{R}^{F \times N}$, $\mathbf{b}_1 \in \mathbb{R}^{N}$ & First-layer weights and biases \\
		$\mathbf{z} = \mathbf{W}_1^\top \mathbf{x} + \mathbf{b}_1$ & First-layer pre-activations \\
		$\mathbf{a} = \relu(\mathbf{z})$ & First-layer activations \\
		$\mathbf{u} \in \mathbb{R}^{K}$ & Logits; $\mathbf{p} = \softmax(\mathbf{u})$ \\
		$\ell(\mathbf{x},y)$   & Per-sample cross-entropy loss \\
		$\mathcal{L}$          & Batch loss $\frac{1}{B}\sum_{i=1}^{B}\ell(\mathbf{x}_i,y_i)$ \\
		$c_{ij} = \partial\mathcal{L}/\partial z_{ij}$ & Sensitivity of sample $i$ at neuron $j$ \\
		$c_s,\ \delta$           & Robust-soliton parameters (Sec.~\ref{sec:bg-lt}) \\
		$\mathbf{G} \in \mathbb{R}^{F \times N}$ & Weight gradient $\nabla_{\mathbf{W}_1} \mathcal{L}$ \\
		$\mathbf{h} \in \mathbb{R}^{N}$ & Bias gradient $\nabla_{\mathbf{b}_1} \mathcal{L}$ \\
		$(\delta\mathbf{G},\delta\mathbf{h})$ & Recovered samples' first-layer gradient (Sec.~\ref{sec:residual}) \\
		$\mathcal{F}$           & Samples certified in the current iteration (Sec.~\ref{sec:residual}) \\
		$\Kset$           & Samples recovered so far (Sec.~\ref{sec:residual}) \\
		$\mathbf{r}[j] \in \mathbb{R}^{F}$ & Candidate vector at neuron $j$, $\mathbf{r}[j] = \mathbf{G}[:,j]/\mathbf{h}[j]$ \\
		$A_j$                  & Activation set $\{i : z_{ij} > 0\}$ \\
		$d_j = |A_j|$          & Degree of neuron $j$ \\
		$N_i$                  & Set of neurons activated by sample $i$ \\
		$s \in (0,1]$          & Trap-weight scale \\
        $p_j$					& firing probability of neuron $j$; $p_j = \frac{d_j}{B}$ \\
		\bottomrule
	\end{tabular}
\end{table}

\subsection{Neural Networks for Classification}
\label{sec:bg-nn}

We consider classification networks that map an input $\mathbf{x} \in \mathbb{R}^{F}$ to logits $\mathbf{u} \in \mathbb{R}^{K}$, with class probabilities $\mathbf{p} = \softmax(\mathbf{u})$. The only architectural assumption required by our attacks concerns the first layer: the input passes through a fully connected layer followed by a ReLU activation,
\begin{equation}
	\mathbf{z} = \mathbf{W}_1^\top \mathbf{x} + \mathbf{b}_1 \in \mathbb{R}^{N}, 
	\quad 
	\mathbf{a} = \relu(\mathbf{z}),
	\label{eq:bg-first}
\end{equation}
where $\mathbf{W}_1 \in \mathbb{R}^{F \times N}$ and $\mathbf{b}_1 \in \mathbb{R}^{N}$. All subsequent layers may be arbitrary. Training uses the cross-entropy loss. For a single sample $(\mathbf{x},y)$,
\begin{equation}
	\ell(\mathbf{x},y) = -\log p_y, 
	\qquad 
	\frac{\partial \ell}{\partial u_k} = p_k - \mathbb{I}[k=y].
	\label{eq:bg-ce}
\end{equation}
For a batch of $B$ samples the loss is the average $\mathcal{L} = \frac{1}{B}\sum_{i=1}^{B}\ell(\mathbf{x}_i,y_i)$, and all gradients considered in this paper are taken with respect to this batch-averaged loss.

\subsection{First-Layer Gradients}
\label{sec:bg-grad}
The attacks rely on the structure of the first-layer gradient. For a single sample $(\mathbf{x},y)$, the first-layer pre-activation and activation of neuron $j$ are
\[
z_j=\mathbf{W}_1[:,j]^\top\mathbf{x}+\mathbf{b}_1[j],
\qquad
a_j=\relu(z_j),
\]
and define the sensitivity of neuron $j$ as $c_j=\partial\ell/\partial z_j$. Since
\[
\frac{\partial a_j}{\partial z_j}
=\mathbb{I}[z_j>0]
\qquad\Longrightarrow\qquad
c_j
=\frac{\partial\ell}{\partial z_j}
=\frac{\partial\ell}{\partial a_j}\,\mathbb{I}[z_j>0],
\]
$c_j=0$ whenever the ReLU is inactive. Applying the chain rule once more yields the per-sample gradients
\begin{align}
	\nabla_{\mathbf{W}_1[:,j]}\ell
	&=\frac{\partial\ell}{\partial z_j}
	  \frac{\partial z_j}{\partial\mathbf{W}_1[:,j]}
	=c_j\,\mathbf{x},\\
	\nabla_{\mathbf{b}_1[j]}\ell
	&=\frac{\partial\ell}{\partial z_j}
	  \frac{\partial z_j}{\partial\mathbf{b}_1[j]}
	=c_j.
	\label{eq:bg-persample}
\end{align}
Consequently, when a neuron is activated by only one sample, the ratio of the corresponding weight and bias gradients recovers that sample exactly:
\begin{equation}
    \frac{\nabla_{\mathbf{W}_1[:,j]}\mathcal{L}}
    {\nabla_{\mathbf{b}_1[j]}\mathcal{L}}
    = \mathbf{x}_i.
    \label{eq:bg-ratio}
\end{equation}
This identity is the basic recovery primitive used by prior analytic attacks~\cite{Boenisch2023Curious} and by the iterative procedure developed in this work. For a batch, the first-layer gradients accumulate the contributions of all activating samples. Writing
\begin{equation}
	\mathbf{G} = \nabla_{\mathbf{W}_1}\mathcal{L} \in \mathbb{R}^{F\times N}, 
	\quad 
	\mathbf{h} = \nabla_{\mathbf{b}_1}\mathcal{L} \in \mathbb{R}^{N},
\end{equation}
the $j$-th column satisfies
\begin{equation}
	\mathbf{G}[:,j] = \sum_{i\in A_j} c_{ij}\,\mathbf{x}_i, 
	\qquad 
	\mathbf{h}[j] = \sum_{i\in A_j} c_{ij},
	\label{eq:bg-rows}
\end{equation}
where $A_j = \{i:z_{ij}>0\}$ is the activation set of neuron $j$ and $c_{ij}=\partial\mathcal{L}/\partial z_{ij}$.

\begin{definition}[Sensitivity, activation set, and degree]
    \label{def:degree}
	The \emph{sensitivity} of sample $i$ at neuron $j$ is $c_{ij} = \partial\mathcal{L}/\partial z_{ij}$.
	The \emph{activation set} of neuron $j$ is $A_j = \{i : z_{ij} > 0\}$, and its \emph{degree} is $d_j = |A_j|$.
\end{definition}

\begin{definition}[Singleton and isolated sample]
	A neuron of degree one is called a \emph{singleton}. The unique sample that activates a singleton is said to be \emph{isolated} at that neuron.
\end{definition}

If sample $i$ is isolated at neuron $j$, then~\eqref{eq:bg-rows} reduces to a single term and the ratio $\mathbf{G}[:,j]/\mathbf{h}[j]$ recovers $\mathbf{x}_i$ exactly.

\subsection{Luby Transform Codes}
\label{sec:bg-lt}

The cascading recovery procedure introduced in Section~\ref{sec:method} is analogous to the peeling decoder of Luby Transform (LT) codes~\cite{Luby2002LT}. In an LT code, a set of $B$ message bits is encoded into a stream of check bits, each formed as the XOR of a randomly chosen subset of the message bits. Decoding proceeds by iteratively locating degree-1 check bits (those with only one message bit), recovering the associated message bit, and subtracting (XORing) its contribution from all neighboring check bits. This operation reduces the degrees of the remaining checks and frequently creates new degree-1 nodes, resulting in a cascade that recovers the entire message with high probability when the check-degree distribution is appropriately chosen.

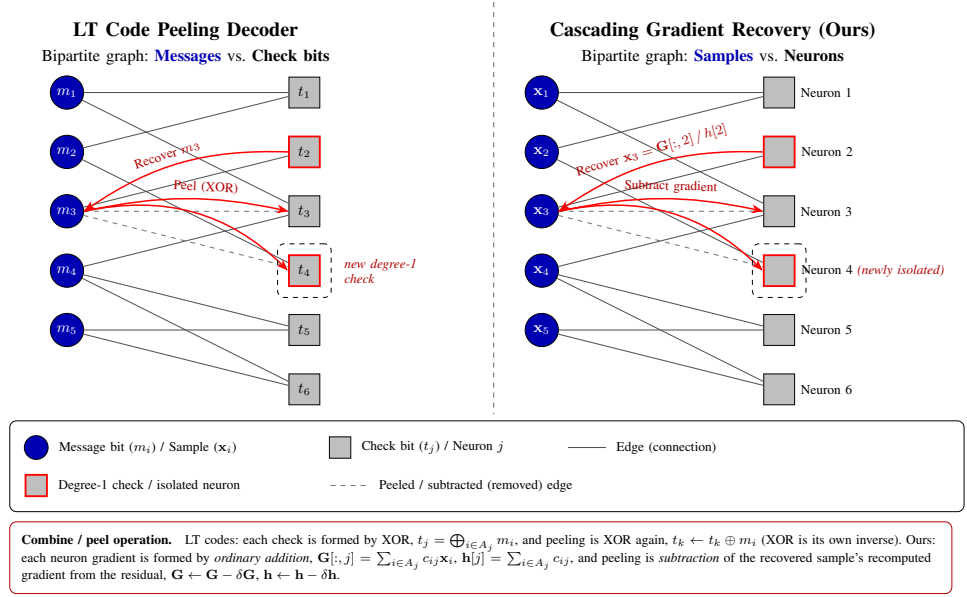
\begin{figure*}[t]
	\centering
	\resizebox{0.7\textwidth}{!}{
		\begin{tikzpicture}[
			font=\small,
			msg/.style={circle, draw, fill=blue!70!black, text=white,
				minimum size=7mm, inner sep=0pt, font=\small\bfseries},
			chk/.style={rectangle, draw, fill=black!25,
				minimum size=6.5mm, inner sep=0pt, font=\small},
			hot/.style={draw=red, line width=1.1pt},
			edge/.style={draw=black!70},
			peeled/.style={draw=black!55, dashed},
			rec/.style={-{Stealth[length=2.5mm]}, draw=red, line width=0.9pt},
			lbl/.style={font=\footnotesize, text=red!70!black},
			panel/.style={font=\large\bfseries},
			sub/.style={font=\normalsize},
			]
			
			\begin{scope}[local bounding box=LT]
				\foreach \i/\y in {1/0, 2/-1.25, 3/-2.5, 4/-3.75, 5/-5.0}
				\node[msg] (m\i) at (0,\y) {$m_\i$};
				
				\foreach \j/\y in {1/0, 3/-2.5, 5/-5.0, 6/-6.25}
				\node[chk] (t\j) at (5,\y) {$t_\j$};
				
				\node[chk, hot] (t2) at (5,-1.25) {$t_2$};
				\node[chk, hot] (t4) at (5,-3.75) {$t_4$};
				
				\draw[edge] (m1) -- (t1);
				\draw[edge] (m2) -- (t1);
				\draw[edge] (m3) -- (t2);
				\draw[edge] (m1) -- (t3);
				\draw[edge] (m4) -- (t3);
				\draw[edge] (m2) -- (t4);
				\draw[edge] (m4) -- (t5);
				\draw[edge] (m5) -- (t5);
				\draw[edge] (m4) -- (t6);
				\draw[edge] (m5) -- (t6);
				
				\draw[peeled] (m3) -- (t3);
				\draw[peeled] (m3) -- (t4);
				
				\draw[rec] (t2.west) to[bend right=18]
				node[lbl, above, sloped, pos=0.55]{Recover $m_3$} (m3.east);
				\draw[rec] (m3.east) to[bend left=12]
				node[lbl, above, sloped, pos=0.6]{Peel (XOR)} (t3.west);
				\draw[rec] (m3.east) to[bend left=28] (t4.west);
				
				\node[draw, dashed, rounded corners, inner sep=2.2mm, fit=(t4)] (newt) {};
				\node[right=1.5mm of newt, font=\footnotesize\itshape, text=red!70!black, align=left]
				{new degree-1\\check};
				
				\node[panel] at (2.5,1.3) {LT Code Peeling Decoder};
				\node[sub] at (2.5,0.75) {Bipartite graph: \textcolor{blue!70!black}{\textbf{Messages}} vs.\ \textbf{Check bits}};
			\end{scope}
			
			\begin{scope}[local bounding box=OURS, xshift=10cm]
				\foreach \i/\y in {1/0, 2/-1.25, 3/-2.5, 4/-3.75, 5/-5.0}
				\node[msg] (x\i) at (0,\y) {$\mathbf{x}_\i$};
				
				\foreach \j/\y in {1/0, 3/-2.5, 5/-5.0, 6/-6.25}
				\node[chk, label={[font=\footnotesize]right:Neuron \j}] (n\j) at (5,\y) {};
				
				\node[chk, hot, label={[font=\footnotesize]right:Neuron 2}] (n2) at (5,-1.25) {};
				\node[chk, hot, label={[font=\footnotesize]right:{Neuron 4 \textcolor{red!70!black}{\itshape(newly isolated)}}}] (n4) at (5,-3.75) {};
				
				\draw[edge] (x1) -- (n1);
				\draw[edge] (x2) -- (n1);
				\draw[edge] (x3) -- (n2);
				\draw[edge] (x1) -- (n3);
				\draw[edge] (x4) -- (n3);
				\draw[edge] (x2) -- (n4);
				\draw[edge] (x4) -- (n5);
				\draw[edge] (x5) -- (n5);
				\draw[edge] (x4) -- (n6);
				\draw[edge] (x5) -- (n6);
				
				\draw[peeled] (x3) -- (n3);
				\draw[peeled] (x3) -- (n4);
				
				\draw[rec] (n2.west) to[bend right=18]
				node[lbl, above, sloped, pos=0.5]{Recover $\mathbf{x}_3 = \mathbf{G}[:,2]\,/\,h[2]$} (x3.east);
				\draw[rec] (x3.east) to[bend left=12]
				node[lbl, above, sloped, pos=0.55]{Subtract gradient} (n3.west);
				\draw[rec] (x3.east) to[bend left=28] (n4.west);
				
				\node[draw, dashed, rounded corners, inner sep=2.2mm, fit=(n4)] (newn) {};
				
				\node[panel] at (3.6,1.3) {Cascading Gradient Recovery (Ours)};
				\node[sub] at (3.6,0.75) {Bipartite graph: \textcolor{blue!70!black}{\textbf{Samples}} vs.\ \textbf{Neurons}};
			\end{scope}
			
			\draw[dashed, black!60] ($(LT.north east)+(1.5,0.3)$) -- ($(LT.south east)+(1.5,-0.2)$);
			
			\begin{scope}[shift={($(LT.south west)+(0,-0.9)$)}]
				\node[msg, minimum size=5mm] (lg1) at (0,0) {};
				\node[right=1mm of lg1, font=\footnotesize, anchor=west] (lg1t)
				{Message bit ($m_i$) / Sample ($\mathbf{x}_i$)};
				\node[chk, minimum size=4.5mm] (lg2) at (6.4,0) {};
				\node[right=1mm of lg2, font=\footnotesize, anchor=west] (lg2t)
				{Check bit ($t_j$) / Neuron $j$};
				\draw[edge] (11.2,0) -- (12.0,0);
				\node[right=1mm, font=\footnotesize, anchor=west] (lg3t) at (12.0,0)
				{Edge (connection)};
				\node[chk, hot, minimum size=4.5mm] (lg4) at (0,-0.8) {};
				\node[right=1mm of lg4, font=\footnotesize, anchor=west] (lg4t)
				{Degree-1 check / isolated neuron};
				\draw[peeled] (6.2,-0.8) -- (7.0,-0.8);
				\node[right=1mm, font=\footnotesize, anchor=west] (lg5t) at (7.0,-0.8)
				{Peeled / subtracted (removed) edge};
				\begin{scope}[on background layer]
					\node[draw, rounded corners, inner sep=3mm,
					fit={(lg1)(lg1t)(lg2t)(lg3t)(lg4)(lg4t)(lg5t)
						($(LT.west |- lg1.north)$)($(OURS.east |- lg5t.south)$)}] (legend) {};
				\end{scope}
			\end{scope}
			
			\path let \p1=(OURS.east), \p2=(LT.west) in
			node[draw=red!70!black, rounded corners, inner sep=2.5mm, anchor=north west,
			font=\footnotesize, align=left, text width={\x1-\x2+1mm},
			below=2mm of legend.south west, anchor=north west] (note)
			{\textbf{Combine / peel operation.}\quad
				LT codes: each check is formed by XOR, $t_j = \bigoplus_{i \in A_j} m_i$,
				and peeling is XOR again, $t_k \leftarrow t_k \oplus m_i$ (XOR is its own inverse).
				Ours: each neuron gradient is formed by \emph{ordinary addition},
				$\mathbf{G}[:,j] = \sum_{i \in A_j} c_{ij} \mathbf{x}_i$, $\mathbf{h}[j] = \sum_{i \in A_j} c_{ij}$,
				and peeling is \emph{subtraction} of the recovered sample's recomputed gradient
				from the residual, $\mathbf{G} \leftarrow \mathbf{G} - \delta \mathbf{G}$, $\mathbf{h} \leftarrow \mathbf{h} - \delta \mathbf{h}$.};
			
	\end{tikzpicture}}
	\caption{Analogy between the LT-code peeling decoder (left) and the cascading gradient recovery process proposed in this work (right). In both cases, degree-1 nodes allow exact recovery of a connected left-hand node. Subtracting (peeling) the recovered node from the residual structure creates new degree-1 nodes, enabling an iterative cascade. In our setting, samples correspond to message bits and neurons correspond to check bits.}
	\label{fig:peeling}
\end{figure*}

In the present setting, the $B$ input samples play the role of message bits, while the $N$ first-layer neurons act as check nodes. A neuron of degree one isolates a single sample, which can be recovered exactly via the ratio of the corresponding weight-gradient and bias-gradient entries. Subtracting the recovered sample’s contribution from the residual gradient is analogous to the peeling step and can produce new isolated neurons, thereby propagating the recovery cascade. This parallel is illustrated in Fig.~\ref{fig:peeling}.

\noindent\textbf{LT Encoding.}

Given $B$ message bits $m_1,\dots,m_B\in\{0,1\}$, each check bit is generated as follows:
\begin{enumerate}
    \item Draw a degree $d$ from a prescribed distribution $\rho(d)$ supported on $\{1,\dots,B\}$.
    \item Select a subset $A_j\subset\{1,\dots,B\}$ of size $d$ uniformly at random.
    \item Form the check bit
    \[
    t_j = \bigoplus_{i\in A_j} m_i.
    \]
\end{enumerate}
The resulting bipartite graph has message nodes on the left and check nodes on the right, with an edge between message $i$ and check $j$ whenever $i\in A_j$. The degree of check node $j$ is $d_j=|A_j|$.

In our gradient-inversion setting the same structure arises: each neuron $j$ is activated by a subset $A_j$ of the batch, and the observed gradient column $(\mathbf{G}[:,j],\mathbf{h}[j])$ encodes a linear combination of the corresponding samples. By controlling the first-layer parameters we can influence the distribution of the degrees $d_j$.

\noindent\textbf{LT Decoding (Peeling).}

Whenever a check node of degree one exists, its check bit equals the unique neighboring message bit. The decoder therefore proceeds as follows:
\begin{enumerate}
    \item Locate a check node $j$ with $d_j=1$ and let $m_i$ be its sole neighbor.
    \item Recover $m_i\leftarrow t_j$.
    \item For every remaining check $k$ adjacent to $m_i$, update
    \[
    t_k\leftarrow t_k\oplus m_i
    \]
    and decrement $d_k$ by one.
    \item Remove the recovered message node and all of its incident edges.
    \item Repeat until either all message bits are recovered or no degree-1 checks remain.
\end{enumerate}
The success of this process is governed by the degree distribution of the check nodes. When the degrees follow the robust soliton distribution, recovery succeeds with high probability provided the number of checks satisfies
\[
N\gtrsim B\bigl(1+\varepsilon\bigr),
\]
where the overhead $\varepsilon=O\bigl(\ln^2(B/\delta)/\sqrt{B}\bigr)$ guarantees a failure probability of at most $\delta$. This guarantee relies on uniformly random edge selection; the extent to which it transfers to our data-dependent setting is discussed in Section~\ref{sec:param_manipulation}.

\noindent\textbf{The Robust Soliton Distribution.}

The ideal soliton distribution on $\{1,\dots,B\}$ is defined by
\begin{align*}
    \rho_{\mathrm{ideal}}(1)&=\frac{1}{B},\\
    \rho_{\mathrm{ideal}}(i)&=\frac{1}{i(i-1)},\qquad i=2,\dots,B.
\end{align*}
The robust soliton distribution is obtained by adding a carefully chosen perturbation $\tau(i)$ and renormalizing. Let
\[
R=c_s\ln\Bigl(\frac{B}{\delta}\Bigr)\sqrt{B},
\]
where $c_s>0$ and $\delta\in(0,1)$ are design parameters. The perturbation is given by
\begin{align*}
    \tau(i)&=\frac{R}{iB},\qquad i=1,\dots,\bigl\lfloor B/R\bigr\rfloor-1,\\
    \tau\bigl(\lfloor B/R\rfloor\bigr)&=\frac{R\ln(R/\delta)}{B},\\
    \tau(i)&=0,\qquad i>\lfloor B/R\rfloor.
\end{align*}
The resulting distribution places additional mass near degree $B/R$, which supplies a reservoir of higher-degree checks that sustain the peeling process after the initial wave of degree-1 recoveries.

In Sections~\ref{sec:sf} and~\ref{sec:sd} we deliberately shape the activation degrees of the first-layer neurons toward this robust soliton distribution, thereby maximizing recovery.

\section{Related Work and Positioning}
\label{sec:related}
	
	We position our attack against prior work in four areas: sparsity-based sample isolation, recovery bounds for isolation-based methods, exact batch reconstruction, and label recovery.

	\paragraph{Sparsity-based analytic attacks.}
	The closest prior work is the attack of Boenisch et al.~\cite{Boenisch2023Curious} (CaH). Their method recovers only isolated samples (singletons) and introduces mirrored trap weights to induce a sparse activation pattern whose degrees approximately follow a Poisson distribution with mean one. Our approach differs in two fundamental respects. First, we continue recovery beyond the initial set of isolated samples through an iterative subtraction process that creates new singletons, forming a recovery cascade. Second, we deliberately shape the neuron degree distribution toward the robust soliton distribution, which Luby~\cite{Luby2002LT} showed to be near-optimal for peeling-style decoding, rather than relying on a Poisson distribution.
	
	\paragraph{Recovery bounds for isolation-based methods.}
	Diana et al.~\cite{diana2025cutting} proved that any attack relying solely on isolated samples is upper-bounded by the number of vertices of the convex hull of the batch. A sample isolated by a neuron corresponds geometrically to a vertex that the hyperplane defined by that neuron’s weights and bias separates from the remaining samples. For data drawn from continuous distributions the resulting bounds grow slowly with batch size:
	\begin{enumerate}
		\item $O\bigl(B^{(F-1)/(F+1)}\bigr)$ for the unit ball,
		\item $O\bigl(\log^{F-1} B\bigr)$ for the unit hypercube,
		\item $O\bigl(\log^{(F-1)/2} B\bigr)$ for the standard Gaussian~\cite{raynaud1970convex}.
	\end{enumerate}
	These bounds become particularly restrictive in low dimension, which led the authors to identify tabular data as a weak point of existing sparsity-based attacks. The cascading procedure developed in this paper is not subject to the same limitation: after outer samples are subtracted, interior points become vertices of the remaining set and can themselves be isolated and recovered. Consequently, our Soliton-Data method recovers HARUS batches in full at sizes up to $512$ (Table~\ref{tab:recall_avg_C5_S99}).
	
	\paragraph{Exact batch reconstruction.}
	The sparsity-based attacks discussed above recover samples in closed form from a single gradient, but only while they are isolated. Diana et al.~\cite{diana2025cutting} recover full batches exactly, but this requires multiple communication rounds with carefully controlled per-sample gradient coefficients. Such control is incompatible with cross-device deployments, where a client may participate in only one round. Their follow-up~\cite{Diana2026NoGuessing} reduces the number of rounds and certifies the isolation of each recovered sample. In the honest-but-curious setting, SPEAR~\cite{Dimitrov2024SPEAR}, which its authors describe as the first algorithm to reconstruct whole batches with $B>1$ exactly, exploits the low-rank structure of gradients together with ReLU-induced sparsity, recovering ImageNet-scale inputs for batch sizes up to approximately 25, and SPEAR++~\cite{Bakarsky2025SPEARpp} improves its scalability. Li et al.~\cite{li2024perfect} reformulate exact input reconstruction as the Hidden Subset Sum Problem. Our attack recovers batches exactly from a single FedSGD round, in both the passive and the active setting, by successive isolation and subtraction rather than multi-round search or optimization, and each recovered sample carries its label and a certificate obtained from the same residual fit, without any additional queries.
	
	\paragraph{Label recovery.}
	Analytic label recovery from gradients began with iDLG~\cite{zhao2020idlg}, which extracts the label of a single sample from the signs of the final-layer gradient. Subsequent batch methods (GradInversion~\cite{Yin2021GradInversion}, RLG~\cite{Dang2021RLG}, iLRG~\cite{Ma2023iLRG}) recover only aggregate label statistics of the batch and operate exclusively on the last layer. In contrast, we recover the label of each individual reconstructed sample from the first-layer bias residual via the logit Jacobian. The same procedure simultaneously serves as a certificate that the recovered vector is a genuine training sample, which is required for the subtraction step of the cascade.

\section{Threat Model and Architectural Assumptions}
\label{sec:threat}

\subsection{Threat Model}
\label{sec:bg-threat}

We consider a standard federated learning setting in which a central server distributes the current global model to a set of clients. Each client computes an update on its private local data and returns the update to the server. We focus on single-round FedSGD, in which the client returns the gradient of the loss evaluated on a local batch of size $B$. The server therefore observes the first-layer gradient components $(\mathbf{G},\mathbf{h})$ defined in Section~\ref{sec:bg-grad}. Unlike FedAvg, the client does not perform multiple local SGD steps before returning an update.

The adversary is the server. We distinguish two threat models:

\begin{itemize}
	\item \textbf{Passive (honest-but-curious).}  
	The server follows the prescribed protocol exactly: it transmits an unmodified model and observes only the gradients (or model updates) returned by the clients. No parameter manipulation is performed.
	
	\item \textbf{Active (malicious).}  
	The server is permitted to modify the weights and biases of the model before distribution, while still observing only the returned gradients. In this work we propose both a weight-manipulation method (the trap-weight construction of Section~\ref{sec:trap}) and bias-manipulation methods; for the latter we introduce two variants, one where the server has access to an auxiliary batch of data (Soliton-Data, Section~\ref{sec:sd}) and one where it does not (Soliton-Free, Section~\ref{sec:sf}).
\end{itemize}

Our passive attacks are evaluated on a standard random initialization of the first-layer weights.

\subsection{Architectural Assumptions}
\label{sec:structure}

The proposed attacks rely on a single architectural property: the network begins with a fully-connected layer followed by a ReLU activation. Formally, for an input sample $\mathbf{x}_i\in\mathbb{R}^F$ we have
\begin{equation}
	\begin{aligned}
		\mathbf{z}_i &= \mathbf{W}_1^\top\mathbf{x}_i + \mathbf{b}_1 \in \mathbb{R}^N, \\
		\mathbf{a}_i &= \relu(\mathbf{z}_i),
	\end{aligned}
	\label{eq:first}
\end{equation}
where $\mathbf{W}_1\in\mathbb{R}^{F\times N}$ and $\mathbf{b}_1\in\mathbb{R}^N$. All subsequent layers may be arbitrary; the classification logits may appear any number of layers downstream.

The remainder of the network influences the attack only through the per-sample, per-neuron sensitivities $c_{ij}=\partial\mathcal{L}/\partial z_{ij}$. Once a candidate sample has been recovered, the server can compute these sensitivities by a standard forward and backward pass through its own copy of the model.

\section{Method}
\label{sec:method}

We refer to the iterative recover-and-subtract procedure developed in this section as \emph{peeling}, or equivalently the \emph{cascade}; both terms denote the same process. Throughout, $(\mathbf{G},\mathbf{h})$ denotes the current \emph{residual} gradient and bias, i.e., what remains of the observed gradient after the contributions of all recovered samples have been subtracted; initially the residual is the observed gradient itself.

\subsection{The Baseline: Singleton Extraction}
\label{sec:baseline}

If neuron $j$ is a singleton in the current residual, i.e., $A_j=\{i\}$ among the unrecovered samples, then~\eqref{eq:bg-rows} contains a single term and the unknown sensitivity coefficient cancels in the ratio
\begin{equation}
	\mathbf{r}[j]
	= \frac{\mathbf{G}[:,j]}{\mathbf{h}[j]}
	= \frac{c_{ij}\,\mathbf{x}_i}{c_{ij}}
	= \mathbf{x}_i.
	\label{eq:baseline}
\end{equation}
If instead $|A_j|\ge 2$, the same ratio returns a linear combination of the activating samples:
\begin{equation}
	\mathbf{r}[j]
	= \frac{\mathbf{G}[:,j]}{\mathbf{h}[j]}
	= \sum_{i\in A_j}\lambda_{ij}\,\mathbf{x}_i,
	\qquad
	\lambda_{ij}
	= \frac{c_{ij}}{\sum_{i'\in A_j}c_{i'j}},
	\label{eq:mixture}
\end{equation}
whose coefficients sum to one but need not be positive, because the sensitivities $c_{ij}$ may differ in sign. The resulting combination generally lies outside the data range and does not correspond to a valid sample. Consequently, recovery is upper-bounded by the number of isolated samples, which are the vertices of the convex hull formed by the batch (see Section~\ref{sec:related}).

\subsection{Label Recovery}
\label{sec:labels}

Subtraction of a recovered sample requires its label $y_i$, which is not observed. The label is recovered from the residual bias. Assume the loss is cross-entropy over softmax logits $\mathbf{u}_i$ (located anywhere downstream), so that
\[
\frac{\partial\mathcal{L}}{\partial u_{ik}}
= \frac{1}{B}\bigl(p_{ik}-\mathbb{I}[k=y_i]\bigr).
\]
Let
\begin{equation}
	\mathbf{J}_i
	= \frac{\partial\mathbf{u}_i}{\partial\mathbf{z}_i}
	\in\mathbb{R}^{K\times N}
	\label{eq:jacobian}
\end{equation}
be the Jacobian of the logits with respect to the first-layer pre-activations. The classification layer may sit arbitrarily far downstream. The server evaluates $\mathbf{J}_i$ for a recovered sample by automatic differentiation on its own copy of the model.

A single forward pass of the recovered sample fixes the intermediate activations. Thereafter, the label-independent combination $\mathbf{J}_i^\top\mathbf{p}_i$ is a vector--Jacobian product, obtained with one backward pass (propagating the cotangent $\mathbf{p}_i$ from the logits). Any individual column $\mathbf{J}_i[:,j]$ is obtained via a forward-mode Jacobian--vector product along $\mathbf{e}_j$. For a two-layer network, $\mathbf{J}_i=\mathbf{W}_2^\top$ is constant and no additional passes are required.

For a neuron $j$ that currently isolates sample $i$, the residual bias entry equals the sensitivity, $\mathbf{h}[j]=c_{ij}$. Applying the chain rule component-wise,
\begin{equation}
	\mathbf{h}[j]
	= c_{ij}
	= \Bigl(\mathbf{J}_i^\top\,\frac{\partial\mathcal{L}}{\partial\mathbf{u}_i}\Bigr)_{\!j}
	= \frac{1}{B}\Bigl[
	\bigl(\mathbf{J}_i^\top\mathbf{p}_i\bigr)_j
	- \mathbf{J}_i[y_i,j]
	\Bigr],
	\label{eq:label}
\end{equation}
where the first term is label-independent and the second is label-dependent. Every term except the unknown label $y_i$ is known. Since~\eqref{eq:label} holds exactly for the true label, the label is recovered by trying each of the $K$ candidate labels and keeping the one that minimizes this residual:
\begin{equation}
	\hat{y}_i
	= \argmin_{\ell\in\{1,\dots,K\}}
	\sum_{j\in S_i}
	\Biggl(
	\frac{1}{B}\Bigl[
	\bigl(\mathbf{J}_i^\top\mathbf{p}_i\bigr)_j
	- \mathbf{J}_i[\ell,j]
	\Bigr]
	- \mathbf{h}[j]
	\Biggr)^2,
	\label{eq:labelfit}
\end{equation}
i.e., the label for which this residual vanishes, where $S_i$ is the set of neurons that currently isolate sample $i$ (those whose candidate $\mathbf{r}[j]$ equals $\hat{\mathbf{x}}_i$). The fit uses only those neurons: the label-independent term $\mathbf{J}_i^\top\mathbf{p}_i$ is computed once per sample, after which each candidate label requires one table lookup per neuron $j\in S_i$. The computational cost is one forward pass, one backward pass, and $|S_i|$ Jacobian–vector products per sample (typically a small number). When $\mathbf{J}_i$ is constant the procedure reduces to pure arithmetic. No per-class backward passes and no auxiliary data are required.

\begin{proposition}[Identifiability]
	Two labels $\ell\neq\ell'$ are indistinguishable under~\eqref{eq:labelfit} only if $\mathbf{J}_i[\ell,j]=\mathbf{J}_i[\ell',j]$ for every neuron $j\in S_i$. For weights drawn from any continuous distribution and $|S_i|\ge 1$ this event has probability zero. Because the fit uses exclusively the neurons that isolate sample $i$, duplicate labels elsewhere in the batch do not interfere.
\end{proposition}

\subsection{Certification: Distinguishing Genuine Candidates}
\label{sec:certify}

The ratio $\mathbf{r}$ (Eq.~\eqref{eq:mixture}) produces one candidate per neuron with a nonzero residual bias entry. As shown by~\eqref{eq:mixture}, some of these candidates are linear combinations rather than true samples. The attacker must distinguish the two without access to ground-truth data, because subtraction is destructive: subtracting an invalid combination injects error into the residual and corrupts every subsequent recovery.

The same consistency check that recovers the label simultaneously serves as a certificate of authenticity. For a candidate $\hat{\mathbf{x}}$, the server simulates a forward and backward pass (possible in closed form because it owns every network parameter) and compares the predicted sensitivity against the observed residual bias on the neurons that produced this candidate:
\begin{equation}
	\varepsilon(\hat{\mathbf{x}},\ell)
	= \Biggl\|
	\frac{1}{B}\Bigl[
	\bigl(\mathbf{J}^\top\hat{\mathbf{p}}\bigr)_{S_{\hat{\mathbf{x}}}}
	- \mathbf{J}[\ell, S_{\hat{\mathbf{x}}}]
	\Bigr]
	- \mathbf{h}[S_{\hat{\mathbf{x}}}]
	\Biggr\|,
	\label{eq:cert}
\end{equation}
where $S_{\hat{\mathbf{x}}}=\{j:\mathbf{r}[j]=\hat{\mathbf{x}}\}$ is the set of neurons whose candidate equals $\hat{\mathbf{x}}$, i.e., the duplicates of $\hat{\mathbf{x}}$ in $\mathbf{r}$. For a genuine sample these are exactly the neurons that currently isolate it, so $S_{\hat{\mathbf{x}}}$ coincides with the set $S_i$ of~\eqref{eq:labelfit} once $\hat{\mathbf{x}}$ is certified as sample $i$. A candidate that is a genuine sample satisfies~\eqref{eq:label} exactly on these neurons. In contrast, a linear combination $\sum_i\lambda_i\mathbf{x}_i$ has no single sample behind it: its simulated activation mask and sensitivities are those of the combination itself, not of any summand, and no label makes $\varepsilon(\hat{\mathbf{x}},\ell)$ small.

\paragraph*{Deduplication.}
A sample is typically isolated by several neurons simultaneously, so the same vector appears multiple times among the candidates $\mathbf{r}[j]$. Before certification, the candidates are compared pairwise and those within a small distance tolerance are merged: one copy is kept and certified once, and the neuron indices of all its duplicates are pooled into its set $S_{\hat{\mathbf{x}}}$. Candidates duplicating an already recovered sample are likewise dropped, and their neurons are recorded as having handed over that sample. Deduplication therefore determines exactly the sets used by~\eqref{eq:labelfit} and~\eqref{eq:cert}, and ensures each distinct sample is certified at most once per iteration.

\subsection{Peeling}
\label{sec:residual}

After forming the ratios, each candidate is certified and, if accepted, is assigned a label. The contribution of every certified sample is then subtracted from the observed batch gradient. This subtraction can create new degree-1 neurons and thereby enable further recoveries, producing a cascade.

In the implementation all samples certified within one iteration are subtracted jointly. Their combined contribution is the gradient that would have been returned on a batch consisting solely of those samples, while the loss remains averaged over the original batch size $B$:
\begin{align}
	(\delta\mathbf{G},\delta\mathbf{h})
	&\gets
	\nabla_{(\mathbf{W}_1,\mathbf{b}_1)}
	\frac{1}{B}\sum_{\mathbf{x}\in\mathcal{F}}\ell(\mathbf{x},\hat{y}_{\mathbf{x}}),
	\label{eq:residual}\\
	\mathbf{G}
	&\gets\mathbf{G}-\delta\mathbf{G},
	\qquad
	\mathbf{h}
	\gets\mathbf{h}-\delta\mathbf{h}.
\end{align}
Here $\mathcal{F}$ denotes the set of samples certified in the current iteration; we call $(\delta\mathbf{G},\delta\mathbf{h})$ the recovered samples' first-layer gradient. It is obtained with one batched backward pass. After subtraction, the residual retains the same analytic form, now taken only over the still-unrecovered samples:
\begin{equation}
	\mathbf{G}[:,j]
	= \sum_{i\in A_j\setminus\mathcal{K}}c_{ij}\,\mathbf{x}_i,
	\qquad
	\mathbf{h}[j]
	= \sum_{i\in A_j\setminus\mathcal{K}}c_{ij},
	\label{eq:residual-col}
\end{equation}
where $\mathcal{K}$ is the set of samples recovered so far. The singleton identity~\eqref{eq:baseline} continues to apply unchanged. A neuron of original degree $k$ yields its $k$-th sample as soon as the other $k-1$ samples have been peeled away. Each successful subtraction can therefore expose new singletons, allowing the recovery process to cascade.

\subsection{The Loop}
\label{sec:loop}

The complete iterative procedure is summarized in Algorithm~\ref{alg:main}.

\begin{algorithm}[t]
	\caption{The Peeling Decoder}
	\label{alg:main}
	\begin{algorithmic}[1]
		\Require First-layer gradient $(\nabla_{\mathbf{W}_1}\mathcal{L},\,\nabla_{\mathbf{b}_1}\mathcal{L})$; server's own parameters; batch size $B$
		\State $\mathbf{G}\gets\nabla_{\mathbf{W}_1}\mathcal{L}$,\quad $\mathbf{h}\gets\nabla_{\mathbf{b}_1}\mathcal{L}$,\quad $\mathcal{K}\gets\emptyset$
		\Repeat
		\State $\mathbf{r}[j]\gets\mathbf{G}[:,j]/\mathbf{h}[j]$ for every neuron $j$
		\Comment{candidates, Eq.~\eqref{eq:baseline}}
		\State Merge duplicates: group the neurons $j$ whose candidates
		$\mathbf{r}[j]$ coincide, keep one candidate per group with pooled neuron set
		$S_{\hat{\mathbf{x}}}$, and drop duplicates of samples in $\mathcal{K}$
		\Comment{deduplication}
		\State $\mathcal{F}\gets$ distinct candidates certified by~\eqref{eq:cert}, each with label $\hat{y}$ from the fit~\eqref{eq:labelfit} against the current $\mathbf{h}$
		\State $(\delta\mathbf{G},\delta\mathbf{h})
		\gets
		\nabla_{(\mathbf{W}_1,\mathbf{b}_1)}
		\frac{1}{B}\sum_{\mathbf{x}\in\mathcal{F}}\ell(\mathbf{x},\hat{y}_{\mathbf{x}})$
		\Comment{one batched backward pass, Eq.~\eqref{eq:residual}}
		\State $\mathbf{G}\gets\mathbf{G}-\delta\mathbf{G}$,\quad
		$\mathbf{h}\gets\mathbf{h}-\delta\mathbf{h}$,\quad
		$\mathcal{K}\gets\mathcal{K}\cup\mathcal{F}$
		\Comment{peel the certified set}
		\Until{no new sample is certified}
	\end{algorithmic}
\end{algorithm}

Each neuron can yield at most one sample, because a neuron becomes a singleton at most once. The process terminates when no remaining neuron isolates a single unrecovered sample. At that point every neuron with a nonzero residual bias entry is activated by at least two unknown samples, the ratio identity~\eqref{eq:baseline} no longer applies, and the residual gradient is final. Complete recovery of the batch occurs when this terminal state is never reached before all $B$ samples have been extracted.

\section{Parameter Manipulation}
\label{sec:param_manipulation}

To increase the length and success probability of the recovery cascade, we deliberately shape the activation degrees of the first-layer neurons. We first review trap-weight constructions that induce sparsity, and then present two methods that target the robust soliton distribution.

\paragraph{A caveat on the LT-code analogy.}
Luby's bound assumes each check selects its edges uniformly at random; in our setting the edges are fixed by the data, since neuron $j$ connects to sample $i$ exactly when $\mathbf{W}_1[:,j]^\top\mathbf{x}_i + \mathbf{b}_1[j] > 0$. A sample inside the convex hull of the unrecovered samples can therefore never be the sole activator of a neuron until peeling shrinks the hull past it. Empirically it works well: Soliton-Data achieves near-complete recovery at batch sizes of several hundred (Table~\ref{tab:recall_avg_C5_S99}).

\subsection{Trap Weights}
\label{sec:trap}

The server constructs each column of $\mathbf{W}_1$ so that activation is rare. The positive entries of a column are scaled by a constant $s\in(0,1]$, making the column sum negative. Consequently, a generic input yields $z_{ij}<0$ and is blocked by the ReLU. Only an input that is sufficiently correlated with the particular sign pattern of column $j$ activates the neuron. The parameter $s$ controls the strength of this asymmetry: as $s\to 1$ the asymmetry weakens and more neurons fire.

We compare two constructions; the entries of a column are correlated in the mirrored construction and independent in the other:

\paragraph{Mirrored construction.}
Following Algorithm~1 of Boenisch et al.~\cite{Boenisch2023Curious}, a half-length vector of negative values is drawn and the positive half is defined as $-s$ times the same vector; both halves are then randomly permuted into place:
\[
\mathbf{z}_- \sim -|\mathcal{N}(0,\sigma^2\mathbf{I}_{F/2})|,
\qquad
\mathbf{z}_+ = -s\,\mathbf{z}_-.
\]

\paragraph{Independent construction.}
Each entry retains its own magnitude and is independently assigned a positive sign with probability $1/2$, scaled by $s$:
\begin{equation}
    \mathbf{W}_1[f,j] = \begin{cases}
        -|v_f| & \text{w.p. } \tfrac12,\\[2pt]
        +s\,|v_f| & \text{w.p. } \tfrac12,
    \end{cases}
    \qquad \mathbf{v} \sim \mathcal{N}(0,\sigma^2 \mathbf{I}_F).
\end{equation}
Both constructions use $\mathbf{b}_1=\mathbf{0}$ and $\sigma=0.5$, and both induce the same marginal distribution on the weight values. Their firing behavior, however, differs substantially.

\paragraph{Firing patterns.}

\begin{figure}[t]
	\centering
	\begin{subfigure}[b]{0.49\columnwidth}
		\centering
		\includegraphics[ width=.6\linewidth, height=0.22\textheight]{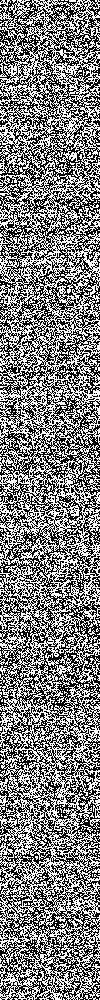}
		\caption{Mirrored weights}
		\label{fig:init-firings-mir}
	\end{subfigure}
	\hfill
	\begin{subfigure}[b]{0.49\columnwidth}
		\centering
		\includegraphics[width=.6\linewidth, height=0.22\textheight]{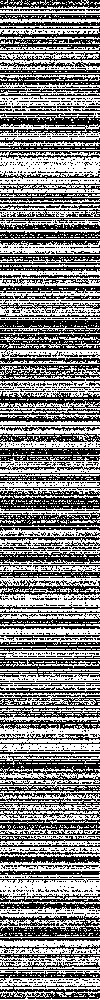}
		\caption{Independent weights}
		\label{fig:init-firings-ind}
	\end{subfigure}
	\caption{Activation patterns (rows: neurons, columns: samples) at $s=1$, $N=1000$, $B=100$. Left: mirrored construction. Right: independent construction. The mirrored pattern lacks the sparse rows that seed the cascade.}
	\label{fig:init-firings}
\end{figure}
\begin{figure}[t]
	\centering
	\includegraphics[width=\linewidth]{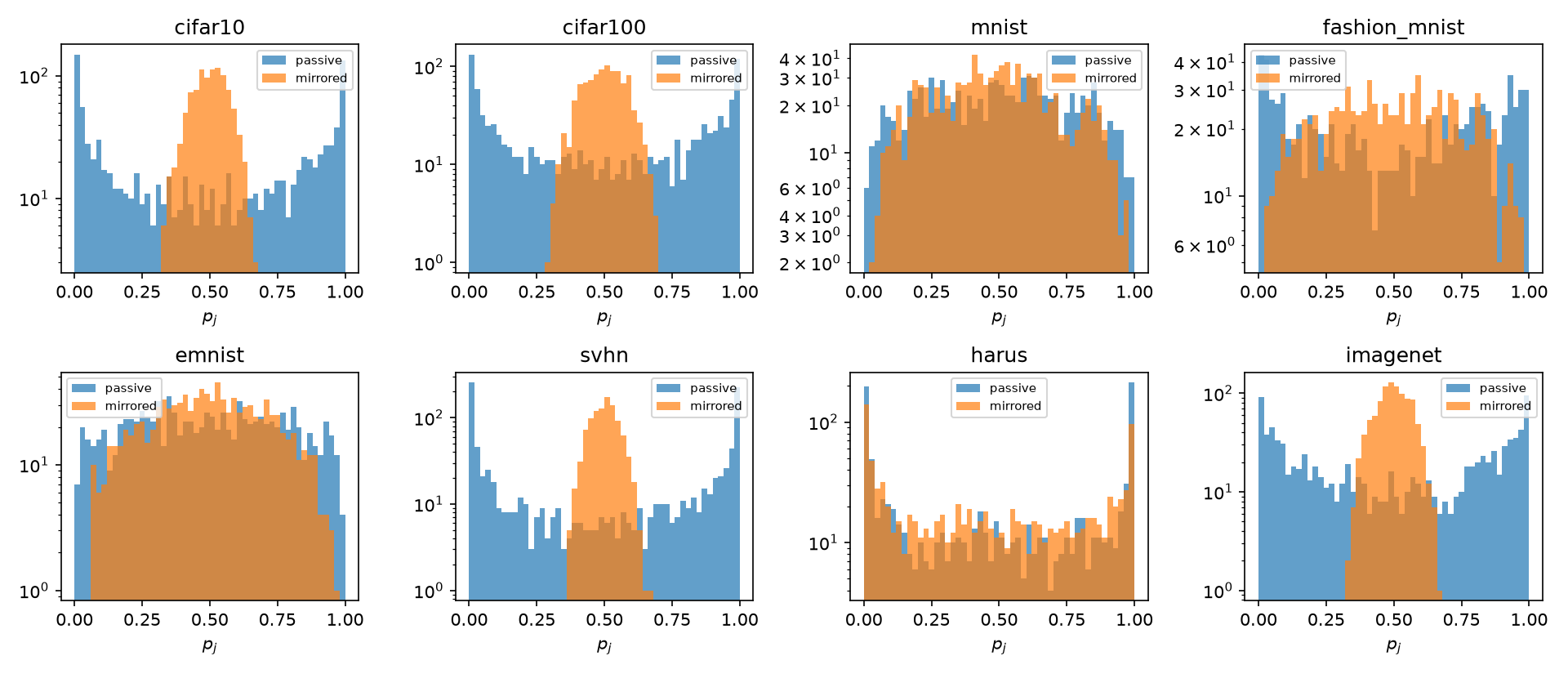}
	\caption{Distribution of neuron firing probabilities $p_j$ under passive (blue) and mirrored (orange) initialization, per data set. Mirroring concentrates $p_j$ near $1/2$ only when the mean image is approximately constant; for MNIST, EMNIST, Fashion-MNIST, and HARUS the distribution stays dispersed.}
	\label{fig:pj-hist}
\end{figure}

Write each sample as $\mathbf{x}_i=\bar{\mathbf{x}}+\boldsymbol{\delta}_i$ (batch mean plus deviation) and each neuron weight as $\mathbf{w}_j=m_j\bar{\mathbf{x}}+\mathbf{u}_j$ with $\mathbf{u}_j\perp\bar{\mathbf{x}}$. The pre-activation then decomposes as
\begin{equation}
	\begin{aligned}
		z_{ij}
		&= \mathbf{w}_j^\top\mathbf{x}_i + \mathbf{b}_1[j]\\
		&= m_j\|\bar{\mathbf{x}}\|^2 + \mathbf{w}_j^\top\boldsymbol{\delta}_i + \mathbf{b}_1[j].
	\end{aligned}
	\label{eq:stripe}
\end{equation}
The resulting activation mask exhibits a stripe pattern: every row possesses its own baseline $m_j\|\bar{\mathbf{x}}\|^2$ determined by the alignment of the neuron with the mean, together with a per-sample deviation term. Neurons with large positive $m_j$ fire on almost every sample (dense rows), while neurons with small or negative $m_j$ fire only on atypical samples whose deviation pushes the pre-activation above zero. These sparse rows are what make singletons possible: if all neurons fired
uniformly, a row of $B$ samples would be a singleton with probability
$B\,2^{-B}$ (roughly $10^{-28}$ at $B=100$) and the cascade could never
begin. That the attack succeeds at all on a randomly initialized network
(Table~\ref{tab:recall_avg_C5_S99}, Passive rows) is direct evidence that
firing is far from uniform, as Fig.~\ref{fig:pj-hist} confirms.

\begin{figure}[t]
	\centering
	\resizebox{0.95\columnwidth}{!}{%
		\includegraphics{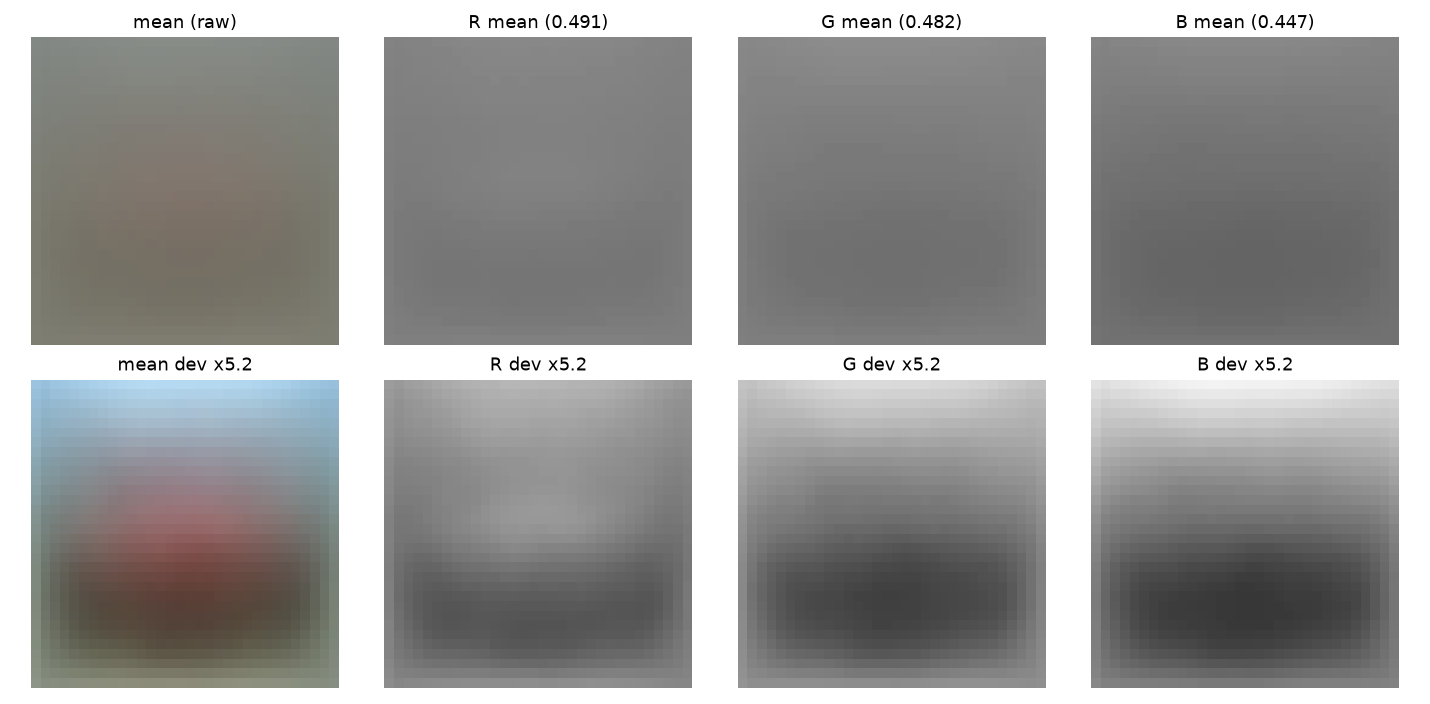}%
	}
	\caption{Left: batch-mean image $\bar{\mathbf{x}}$ (a blurred average CIFAR-10 picture), which constitutes the dominant term in Eq.~\eqref{eq:stripe}. Right: per-channel mean deviations (amplified by $\times 5.2$). Neurons aligned with $\bar{\mathbf{x}}$ fire on nearly all samples, while neurons aligned against it fire only on atypical samples; the latter sparse rows are the seeds of the recovery cascade.}
	\label{fig:means}
\end{figure}

\paragraph{Limitation of the mirrored construction.}
At $s=1$ the mirrored construction matches the marginal distribution of a
standard Gaussian initialization, but it forces every column to sum to
approximately zero. For data sets whose features concentrate about a common
value $a$ (Figure~\ref{fig:means}), this gives
$\mathbf{w}_j^\top\bar{\mathbf{x}}\approx a\sum_f \mathbf{W}_1[f,j]\approx 0$:
the baseline term in~\eqref{eq:stripe} vanishes, the stripe structure and its
sparse bands disappear (Figure~\ref{fig:init-firings}), and recovery collapses
to nearly zero, as Boenisch et al.\ also observe. Figure~\ref{fig:pj-hist} confirms this: mirroring drives the $p_j$ toward
uniform exactly for the data sets whose mean features concentrate around a
single value, which by the argument above eliminates singletons.

\paragraph{Consequence for the passive setting.}
The independent construction at $s=1$ coincides exactly with a standard $\mathcal{N}(0,\sigma^2)$ initialization. Setting $s=1$ therefore provides a meaningful no-tampering reference that corresponds to the honest-but-curious threat model.

\subsection{Soliton-Free (SF)}
\label{sec:sf}

Our goal is to make the neuron degrees $d_j=|A_j|$ follow a robust soliton distribution. Because the server has no access to client data, we adopt a Gaussian approximation based on mild independence assumptions. Features are assumed to be approximately uniform on the unit interval
$[0,1]$ and weakly dependent, so that $\mathbb{E}[x_f]\approx 1/2$,
$\mathrm{Var}[x_f]\approx 1/12$, and covariances are negligible. Under these conditions the pre-activation
\[
z_j = \mathbf{w}_j^\top\mathbf{x} = \sum_{f=1}^F \mathbf{W}_1[f,j]\,x_f
\]
is approximately Gaussian:
\[
z_j\sim\mathcal{N}(\mu_j,\sigma_j^2),
\]
with
\begin{align*}
	\mu_j &= \mathbb{E}[z_j] = \tfrac12\sum_{f=1}^F \mathbf{W}_1[f,j],\\
	\sigma_j^2 &= \mathrm{Var}[z_j] = \tfrac1{12}\sum_{f=1}^F \mathbf{W}_1[f,j]^2.
\end{align*}
Neuron $j$ fires when $z_j+\mathbf{b}_1[j]>0$, i.e., when $z_j>-\mathbf{b}_1[j]$. To ensure that the neuron activates on average for a prescribed degree $d_j$ out of $B$ samples, the threshold $-\mathbf{b}_1[j]$ is placed at the $(1-d_j/B)$-quantile of the Gaussian:
\[
\Pr(z_j>-\mathbf{b}_1[j])=\frac{d_j}{B}.
\]
Solving with the inverse CDF $\Phi^{-1}$ of the standard normal distribution yields
\begin{equation}
	\mathbf{b}_1[j] = -\mu_j - \sigma_j\cdot\Phi^{-1}\Bigl(1-\frac{d_j}{B}\Bigr).
\end{equation}
The degrees $d_j$ themselves are drawn from the robust soliton distribution, thereby shaping the activation pattern toward the distribution that maximizes the success probability of peeling.

\subsection{Soliton-Data (SD)}
\label{sec:sd}

Soliton-Data likewise targets the robust soliton distribution, but exploits a single auxiliary batch to set the degrees exactly rather than through a Gaussian approximation. For each neuron $j$ the server evaluates the pre-activations $\mathbf{w}_j^\top\mathbf{x}$ on the auxiliary samples and sorts them in decreasing order:
\[
\mathbf{w}_j^\top\mathbf{x}_{(1)}\ge\mathbf{w}_j^\top\mathbf{x}_{(2)}\ge\cdots\ge\mathbf{w}_j^\top\mathbf{x}_{(B_{\mathrm{aux}})}.
\]
Because neuron $j$ fires on an input $\mathbf{x}$ precisely when $\mathbf{w}_j^\top\mathbf{x}>-\mathbf{b}_1[j]$, placing the threshold midway between the $d_j$-th and $(d_j+1)$-th largest values,
\begin{equation}
	\mathbf{b}_1[j] = -\frac{\mathbf{w}_j^\top\mathbf{x}_{(d_j)}+\mathbf{w}_j^\top\mathbf{x}_{(d_j+1)}}{2},
\end{equation}
makes the activation set on the auxiliary batch exactly the $d_j$ samples with the largest pre-activations. The auxiliary batch need not come from the clients: in our experiments it is drawn from the test split of each data set, which shares no samples with the clients’ training data. When the auxiliary batch is statistically similar to the clients’ data, the resulting degree distribution on the true batches closely matches the target robust soliton distribution.

\begin{figure*}[t]
	\centering
	\includegraphics[width=0.9\textwidth]{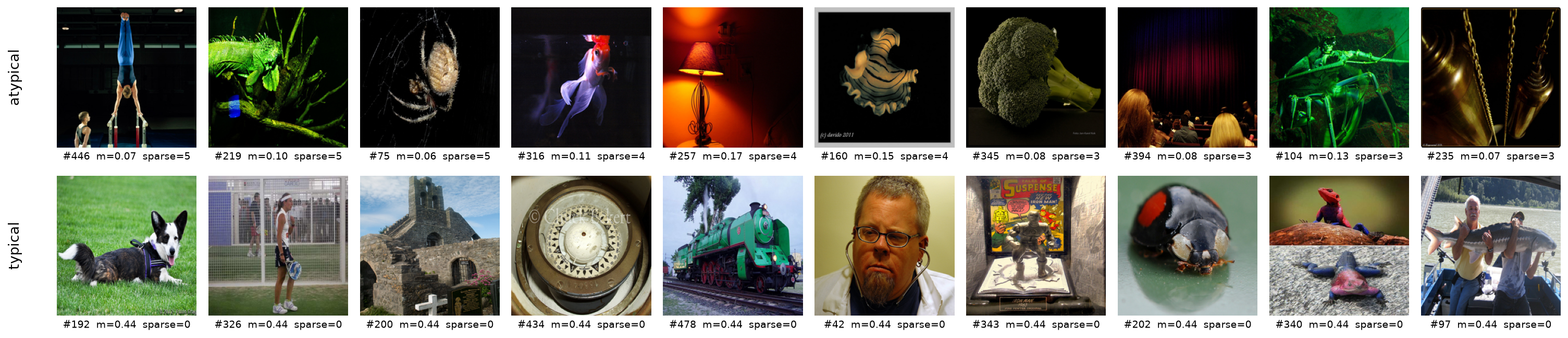}
	\caption{Top: samples that activate sparse neurons (neurons with degree less than 4). Bottom: typical batch samples that the cascade reaches late or never. The atypical samples are visibly off-distribution (darker, atypical ImageNet images). Results shown for $N=1024$, $B=512$, random initialization. $m$ shows the mean pixel value of each image.}
	\label{fig:seeds}
\end{figure*}

\section{Computational Cost}
\label{sec:cost}

Each iteration of the recovery procedure forms all candidate ratios $\mathbf{r}[j]$ in $O(FN)$ time. Duplicate candidates are then merged by pairwise comparison, so that each distinct sample is certified once and its duplicates contribute their neurons to its set $S_{\hat{\mathbf{x}}}$. Certification of each surviving candidate requires one forward pass and one backward pass (the vector--Jacobian product $\mathbf{J}_i^\top\mathbf{p}_i$) performed jointly across all candidates, together with one forward-mode Jacobian--vector product per singleton neuron. In practice, candidates lying outside the data range or possessing a near-zero bias entry are discarded before certification to improve efficiency.

The label fit~\eqref{eq:labelfit} evaluates $K$ candidate labels per candidate, and its residual is the certificate of Section~\ref{sec:certify}; after the $\mathbf{J}_i^\top\mathbf{p}_i$ term has been computed, it reduces to one table lookup per label $\ell$ and per neuron $j\in S_i$ (the neurons whose candidate $\mathbf{r}[j]$ equals $\hat{\mathbf{x}}_i$). Subtraction is performed jointly for all samples certified in the same iteration (Section~\ref{sec:residual}): a single backward pass evaluates the gradient of the combined loss
\[
\frac{1}{B}\sum_{\mathbf{x}\in\mathcal{F}}\ell(\mathbf{x},\hat{y}_{\mathbf{x}}).
\]
Consequently, the number of backward passes equals the number of iterations rather than the number of recovered samples. Memory consumption is dominated by the residual gradient matrix of size $O(FN)$.

\section{Experiments}
\label{sec:experiments}

\paragraph{Experimental setup.}
We evaluate a two-layer multilayer perceptron with $N=1000$ ReLU neurons and softmax cross-entropy loss on eight standard benchmark data sets: CIFAR-10 and CIFAR-100~\cite{Krizhevsky2009CIFAR}, MNIST~\cite{Deng2012MNIST}, EMNIST~\cite{Cohen2017EMNIST}, Fashion-MNIST~\cite{Xiao2017FashionMNIST}, SVHN~\cite{Goodfellow2013SVHN}, ImageNet~\cite{Deng2009ImageNet}, and HARUS~\cite{ReyesOrtiz2013HARUS}. Federated learning is simulated by a single FedSGD round on a batch of size $B$. All experiments are implemented in TensorFlow 2.21.0 and executed in 64-bit floating-point arithmetic on an NVIDIA GeForce RTX 4090 GPU paired with an Intel Core i9-14900K CPU.  First-layer weights use $\sigma=0.5$, second-layer weights use Xavier-uniform initialization~\cite{Glorot2010Understanding}, and all biases are zero except where the soliton constructions set $\mathbf{b}_1$.
The code is publicly available.\footnote{\url{https://github.com/SaeedShariati/Gradient-Inversion-via-LT-Codes}}
\subsection{Recovery Rate}
\label{sec:recall}

Table~\ref{tab:recall_avg_C5_S99} reports the average recovery rate across random seeds for eight data sets. The robust-soliton parameters $(c_s,\delta)$ are used only by the Soliton-Free and Soliton-Data variants; the trap-weight scale $s=0.99$ is used only by the trap-weight baselines. For Soliton-Free and Soliton-Data we set $s=1$. Each method is evaluated under both the independent and the mirrored weight constructions. Note that the trap-weight baselines are strengthened relative to the original CaH attack~\cite{Boenisch2023Curious}: they run the full iterative peeling procedure with certification.

\begin{table*}[!p]
    \centering
    \caption{Average recovery rate across seeds (peeling attack, certificate-admitted) for $(c_s,\delta)=(0.05,0.4)$, $s=0.99$, $N=1000$. Bold indicates recovery rate $>0.50$.}
    \label{tab:recall_avg_C5_S99}
    \scriptsize
    \setlength{\tabcolsep}{2.5pt}
    \renewcommand{\arraystretch}{0.85}
    \resizebox{0.75\linewidth}{!}{%
    \begin{tabular}{@{}ll*{12}{c}@{}}
        \toprule
        \textbf{Dataset} & \textbf{Method} & \textbf{64} & \textbf{128} & \textbf{256} & \textbf{300} & \textbf{350} & \textbf{400} & \textbf{512} & \textbf{600} & \textbf{700} & \textbf{800} & \textbf{900} & \textbf{1024} \\
        \midrule
        \multirow{7}{*}{\textbf{CIFAR-10}} & SF-Ind & \textbf{1.000} & \textbf{1.000} & \textbf{0.905} & \textbf{0.780} & \textbf{0.691} & \textbf{0.630} & \textbf{0.503} & 0.457 & 0.398 & 0.335 & 0.285 & 0.246 \\
        & SF-Mir & \textbf{1.000} & \textbf{1.000} & \textbf{0.711} & \textbf{0.601} & \textbf{0.539} & 0.490 & 0.392 & 0.318 & 0.269 & 0.237 & 0.206 & 0.188 \\
        \cmidrule[0.3pt](lr){2-14}
        & SD-Ind & \textbf{1.000} & \textbf{1.000} & \textbf{0.999} & \textbf{0.997} & \textbf{0.987} & \textbf{0.866} & \textbf{0.755} & \textbf{0.690} & \textbf{0.628} & \textbf{0.580} & \textbf{0.517} & 0.469 \\
        & SD-Mir & \textbf{1.000} & \textbf{1.000} & \textbf{1.000} & \textbf{0.993} & \textbf{0.985} & \textbf{0.876} & \textbf{0.741} & \textbf{0.674} & \textbf{0.620} & \textbf{0.584} & \textbf{0.539} & 0.473 \\
        \cmidrule[0.3pt](lr){2-14}
        & TW-Ind & \textbf{0.962} & \textbf{0.816} & 0.405 & 0.336 & 0.226 & 0.165 & 0.102 & 0.085 & 0.067 & 0.054 & 0.043 & 0.037 \\
        & TW-Mir & 0.000 & 0.000 & 0.000 & 0.000 & 0.000 & 0.000 & 0.000 & 0.000 & 0.000 & 0.000 & 0.000 & 0.000 \\
        \cmidrule[0.3pt](lr){2-14}
        & Passive & \textbf{0.994} & \textbf{0.666} & 0.215 & 0.183 & 0.130 & 0.098 & 0.070 & 0.055 & 0.040 & 0.038 & 0.033 & 0.027 \\
        \midrule
        \multirow{7}{*}{\textbf{CIFAR-100}} & SF-Ind & \textbf{1.000} & \textbf{1.000} & \textbf{0.984} & \textbf{0.875} & \textbf{0.729} & \textbf{0.677} & 0.500 & 0.457 & 0.408 & 0.350 & 0.271 & 0.226 \\
        & SF-Mir & \textbf{1.000} & \textbf{1.000} & \textbf{0.700} & \textbf{0.600} & \textbf{0.520} & 0.460 & 0.376 & 0.325 & 0.288 & 0.250 & 0.225 & 0.195 \\
        \cmidrule[0.3pt](lr){2-14}
        & SD-Ind & \textbf{1.000} & \textbf{1.000} & \textbf{0.999} & \textbf{0.986} & \textbf{0.915} & \textbf{0.812} & \textbf{0.691} & \textbf{0.611} & \textbf{0.545} & 0.487 & 0.465 & 0.411 \\
        & SD-Mir & \textbf{1.000} & \textbf{1.000} & \textbf{0.998} & \textbf{0.969} & \textbf{0.887} & \textbf{0.825} & \textbf{0.689} & \textbf{0.620} & \textbf{0.567} & \textbf{0.510} & 0.468 & 0.417 \\
        \cmidrule[0.3pt](lr){2-14}
        & TW-Ind & \textbf{1.000} & \textbf{1.000} & \textbf{0.576} & 0.319 & 0.213 & 0.155 & 0.095 & 0.078 & 0.070 & 0.060 & 0.047 & 0.038 \\
        & TW-Mir & 0.000 & 0.000 & 0.000 & 0.000 & 0.000 & 0.000 & 0.000 & 0.000 & 0.000 & 0.000 & 0.000 & 0.000 \\
        \cmidrule[0.3pt](lr){2-14}
        & Passive & \textbf{1.000} & \textbf{1.000} & 0.247 & 0.205 & 0.149 & 0.114 & 0.082 & 0.055 & 0.051 & 0.041 & 0.029 & 0.019 \\
        \midrule
        \multirow{7}{*}{\textbf{EMNIST}} & SF-Ind & \textbf{1.000} & \textbf{1.000} & \textbf{0.998} & \textbf{0.567} & 0.341 & 0.266 & 0.193 & 0.156 & 0.128 & 0.107 & 0.096 & 0.080 \\
        & SF-Mir & \textbf{1.000} & \textbf{1.000} & \textbf{1.000} & \textbf{1.000} & \textbf{1.000} & \textbf{0.983} & \textbf{0.518} & 0.317 & 0.251 & 0.222 & 0.187 & 0.148 \\
        \cmidrule[0.3pt](lr){2-14}
        & SD-Ind & \textbf{1.000} & \textbf{1.000} & \textbf{1.000} & \textbf{0.999} & \textbf{0.998} & \textbf{0.997} & \textbf{0.981} & \textbf{0.968} & \textbf{0.913} & \textbf{0.712} & 0.471 & 0.322 \\
        & SD-Mir & \textbf{1.000} & \textbf{1.000} & \textbf{1.000} & \textbf{1.000} & \textbf{0.993} & \textbf{0.986} & \textbf{0.966} & \textbf{0.956} & \textbf{0.911} & \textbf{0.674} & 0.451 & 0.346 \\
        \cmidrule[0.3pt](lr){2-14}
        & TW-Ind & \textbf{0.503} & 0.030 & 0.005 & 0.003 & 0.002 & 0.001 & 0.000 & 0.000 & 0.000 & 0.000 & 0.000 & 0.000 \\
        & TW-Mir & 0.031 & 0.003 & 0.000 & 0.000 & 0.000 & 0.000 & 0.000 & 0.000 & 0.000 & 0.000 & 0.000 & 0.000 \\
        \cmidrule[0.3pt](lr){2-14}
        & Passive & \textbf{0.672} & 0.019 & 0.005 & 0.003 & 0.002 & 0.001 & 0.000 & 0.000 & 0.000 & 0.000 & 0.000 & 0.000 \\
        \midrule
        \multirow{7}{*}{\textbf{Fashion-MNIST}} & SF-Ind & \textbf{1.000} & \textbf{1.000} & \textbf{1.000} & \textbf{0.992} & \textbf{0.749} & 0.417 & 0.307 & 0.234 & 0.181 & 0.140 & 0.125 & 0.109 \\
        & SF-Mir & \textbf{1.000} & \textbf{1.000} & \textbf{1.000} & \textbf{0.987} & \textbf{0.961} & \textbf{0.764} & 0.486 & 0.290 & 0.224 & 0.173 & 0.152 & 0.126 \\
        \cmidrule[0.3pt](lr){2-14}
        & SD-Ind & \textbf{1.000} & \textbf{1.000} & \textbf{1.000} & \textbf{1.000} & \textbf{0.976} & \textbf{0.991} & \textbf{0.925} & \textbf{0.792} & \textbf{0.695} & \textbf{0.582} & \textbf{0.501} & 0.437 \\
        & SD-Mir & \textbf{1.000} & \textbf{1.000} & \textbf{1.000} & \textbf{1.000} & \textbf{1.000} & \textbf{0.973} & \textbf{0.870} & \textbf{0.710} & \textbf{0.666} & \textbf{0.601} & \textbf{0.546} & 0.466 \\
        \cmidrule[0.3pt](lr){2-14}
        & TW-Ind & \textbf{1.000} & \textbf{0.997} & 0.061 & 0.026 & 0.014 & 0.012 & 0.008 & 0.005 & 0.002 & 0.001 & 0.001 & 0.001 \\
        & TW-Mir & 0.469 & 0.017 & 0.001 & 0.001 & 0.001 & 0.001 & 0.000 & 0.000 & 0.000 & 0.000 & 0.000 & 0.000 \\
        \cmidrule[0.3pt](lr){2-14}
        & Passive & \textbf{1.000} & \textbf{0.722} & 0.048 & 0.016 & 0.012 & 0.010 & 0.009 & 0.006 & 0.002 & 0.002 & 0.001 & 0.001 \\
        \midrule
        \multirow{7}{*}{\textbf{HARUS}} & SF-Ind & \textbf{1.000} & 0.409 & 0.120 & 0.110 & 0.087 & 0.064 & 0.045 & 0.038 & 0.031 & 0.029 & 0.024 & 0.018 \\
        & SF-Mir & \textbf{1.000} & 0.358 & 0.127 & 0.091 & 0.069 & 0.058 & 0.045 & 0.038 & 0.028 & 0.024 & 0.021 & 0.020 \\
        \cmidrule[0.3pt](lr){2-14}
        & SD-Ind & \textbf{1.000} & \textbf{1.000} & \textbf{1.000} & \textbf{1.000} & \textbf{1.000} & \textbf{1.000} & \textbf{1.000} & \textbf{0.804} & \textbf{0.601} & \textbf{0.525} & 0.459 & 0.347 \\
        & SD-Mir & \textbf{1.000} & \textbf{1.000} & \textbf{1.000} & \textbf{1.000} & \textbf{1.000} & \textbf{1.000} & \textbf{1.000} & \textbf{0.828} & \textbf{0.690} & \textbf{0.580} & 0.454 & 0.383 \\
        \cmidrule[0.3pt](lr){2-14}
        & TW-Ind & \textbf{1.000} & \textbf{0.603} & 0.147 & 0.111 & 0.084 & 0.070 & 0.054 & 0.046 & 0.041 & 0.034 & 0.030 & 0.023 \\
        & TW-Mir & \textbf{1.000} & \textbf{1.000} & 0.186 & 0.158 & 0.121 & 0.100 & 0.071 & 0.057 & 0.046 & 0.034 & 0.027 & 0.020 \\
        \cmidrule[0.3pt](lr){2-14}
        & Passive & \textbf{1.000} & 0.475 & 0.115 & 0.091 & 0.074 & 0.060 & 0.046 & 0.035 & 0.030 & 0.024 & 0.022 & 0.020 \\
        \midrule
        \multirow{7}{*}{\textbf{ImageNet}} & SF-Ind & \textbf{1.000} & \textbf{1.000} & \textbf{0.993} & \textbf{0.980} & \textbf{0.876} & \textbf{0.817} & \textbf{0.669} & \textbf{0.589} & 0.471 & 0.396 & 0.332 & 0.287 \\
        & SF-Mir & \textbf{1.000} & \textbf{1.000} & \textbf{0.960} & \textbf{0.828} & \textbf{0.723} & \textbf{0.617} & \textbf{0.521} & 0.462 & 0.401 & 0.363 & 0.325 & 0.296 \\
        \cmidrule[0.3pt](lr){2-14}
        & SD-Ind & \textbf{1.000} & \textbf{1.000} & \textbf{0.995} & \textbf{0.973} & \textbf{0.925} & \textbf{0.896} & \textbf{0.744} & \textbf{0.657} & \textbf{0.629} & \textbf{0.538} & 0.495 & 0.448 \\
        & SD-Mir & \textbf{1.000} & \textbf{1.000} & \textbf{0.995} & \textbf{0.992} & \textbf{0.970} & \textbf{0.841} & \textbf{0.764} & \textbf{0.651} & \textbf{0.615} & \textbf{0.576} & \textbf{0.523} & 0.425 \\
        \cmidrule[0.3pt](lr){2-14}
        & TW-Ind & \textbf{1.000} & \textbf{1.000} & \textbf{0.838} & \textbf{0.718} & \textbf{0.608} & 0.474 & 0.413 & 0.311 & 0.245 & 0.199 & 0.146 & 0.114 \\
        & TW-Mir & \textbf{0.612} & \textbf{0.617} & \textbf{0.616} & \textbf{0.613} & \textbf{0.614} & \textbf{0.604} & \textbf{0.593} & 0.498 & 0.341 & 0.016 & 0.004 & 0.001 \\
        \cmidrule[0.3pt](lr){2-14}
        & Passive & \textbf{0.997} & \textbf{0.947} & 0.175 & 0.119 & 0.090 & 0.068 & 0.039 & 0.019 & 0.015 & 0.012 & 0.010 & 0.007 \\
        \midrule
        \multirow{7}{*}{\textbf{MNIST}} & SF-Ind & \textbf{1.000} & \textbf{1.000} & \textbf{0.916} & \textbf{0.537} & 0.357 & 0.271 & 0.174 & 0.141 & 0.114 & 0.094 & 0.078 & 0.066 \\
        & SF-Mir & \textbf{1.000} & \textbf{1.000} & \textbf{0.999} & \textbf{0.985} & \textbf{0.975} & \textbf{0.963} & 0.473 & 0.293 & 0.236 & 0.192 & 0.147 & 0.125 \\
        \cmidrule[0.3pt](lr){2-14}
        & SD-Ind & \textbf{1.000} & \textbf{1.000} & \textbf{1.000} & \textbf{1.000} & \textbf{1.000} & \textbf{1.000} & \textbf{0.978} & \textbf{0.956} & \textbf{0.785} & \textbf{0.637} & \textbf{0.529} & 0.393 \\
        & SD-Mir & \textbf{1.000} & \textbf{1.000} & \textbf{1.000} & \textbf{1.000} & \textbf{1.000} & \textbf{0.995} & \textbf{0.975} & \textbf{0.913} & \textbf{0.799} & \textbf{0.695} & \textbf{0.608} & 0.481 \\
        \cmidrule[0.3pt](lr){2-14}
        & TW-Ind & 0.362 & 0.022 & 0.004 & 0.003 & 0.002 & 0.002 & 0.001 & 0.001 & 0.000 & 0.000 & 0.000 & 0.000 \\
        & TW-Mir & 0.031 & 0.006 & 0.000 & 0.000 & 0.000 & 0.000 & 0.000 & 0.000 & 0.000 & 0.000 & 0.000 & 0.000 \\
        \cmidrule[0.3pt](lr){2-14}
        & Passive & 0.484 & 0.017 & 0.002 & 0.001 & 0.001 & 0.001 & 0.001 & 0.000 & 0.000 & 0.000 & 0.000 & 0.000 \\
        \midrule
        \multirow{7}{*}{\textbf{SVHN}} & SF-Ind & \textbf{1.000} & \textbf{0.925} & 0.405 & 0.308 & 0.285 & 0.244 & 0.190 & 0.165 & 0.137 & 0.122 & 0.088 & 0.075 \\
        & SF-Mir & \textbf{0.869} & 0.394 & 0.206 & 0.167 & 0.142 & 0.127 & 0.087 & 0.085 & 0.071 & 0.058 & 0.051 & 0.042 \\
        \cmidrule[0.3pt](lr){2-14}
        & SD-Ind & \textbf{1.000} & \textbf{1.000} & \textbf{0.856} & \textbf{0.777} & \textbf{0.645} & \textbf{0.557} & 0.492 & 0.419 & 0.395 & 0.347 & 0.319 & 0.289 \\
        & SD-Mir & \textbf{1.000} & \textbf{1.000} & \textbf{0.798} & \textbf{0.686} & \textbf{0.607} & \textbf{0.574} & 0.486 & 0.425 & 0.378 & 0.340 & 0.313 & 0.298 \\
        \cmidrule[0.3pt](lr){2-14}
        & TW-Ind & \textbf{0.894} & \textbf{0.922} & 0.330 & 0.218 & 0.178 & 0.188 & 0.144 & 0.100 & 0.091 & 0.069 & 0.054 & 0.047 \\
        & TW-Mir & 0.006 & 0.000 & 0.000 & 0.000 & 0.000 & 0.000 & 0.000 & 0.000 & 0.000 & 0.000 & 0.000 & 0.000 \\
        \cmidrule[0.3pt](lr){2-14}
        & Passive & \textbf{0.959} & \textbf{0.881} & 0.268 & 0.177 & 0.147 & 0.113 & 0.086 & 0.079 & 0.063 & 0.057 & 0.047 & 0.040 \\
        \bottomrule
    \end{tabular}
    }
\end{table*}

Soliton-Data consistently achieves the highest recovery rate, followed by Soliton-Free. In the passive setting the attack is weakest on MNIST and EMNIST, yet still recovers the majority of samples for batch sizes up to $128$ on CIFAR-100, ImageNet, and several other data sets.

\subsection{Which Samples Are Most at Risk}
\label{sec:grid}

The entire cascade can be visualized on a single binary activation grid of size $N\times B$, where an entry $(j,i)$ is marked if and only if $z_{ij}>0$. Each column of the weight gradient is a linear combination of precisely the samples that activate the corresponding neuron (Eq.~\eqref{eq:bg-rows}). Consequently the singleton rows of the grid constitute the baseline attack, and every subtraction removes one column.

\begin{figure*}[!t]
    \centering
    \resizebox{0.9\linewidth}{!}{%
        \setlength{\fboxrule}{0.15pt}\setlength{\fboxsep}{0pt}%
        \begin{subfigure}[t]{0.15\textwidth}
            {\color{red}\fbox{\includegraphics[height=0.43\textheight]{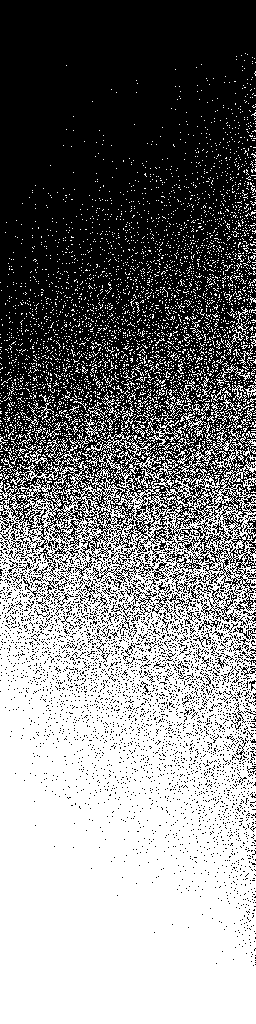}}}
            \caption{Iteration 0: raw grid}
        \end{subfigure}\hfill
        \begin{subfigure}[t]{0.15\textwidth}
            {\color{red}\fbox{\includegraphics[height=0.43\textheight]{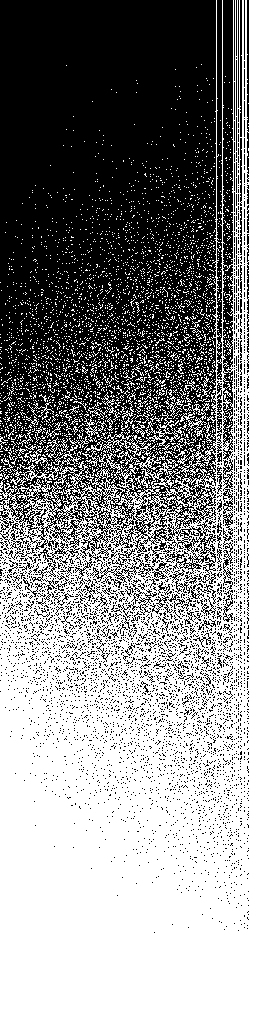}}}
            \caption{Iteration 1: 17 samples removed}
        \end{subfigure}\hfill
        \begin{subfigure}[t]{0.15\textwidth}
            {\color{red}\fbox{\includegraphics[height=0.43\textheight]{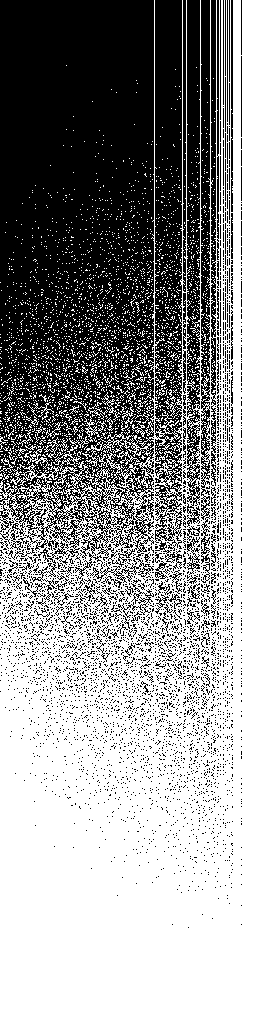}}}
            \caption{Iteration 2: 18 more}
        \end{subfigure}\hfill
        \begin{subfigure}[t]{0.15\textwidth}
            {\color{red}\fbox{\includegraphics[height=0.43\textheight]{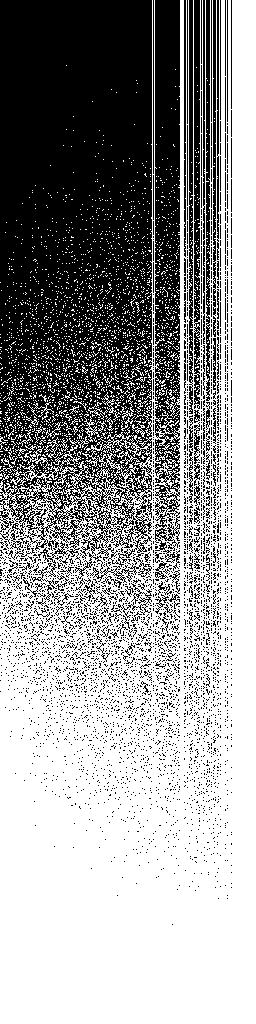}}}
            \caption{Iteration 3: 13 more}
        \end{subfigure}\hfill
        \begin{subfigure}[t]{0.15\textwidth}
            {\color{red}\fbox{\includegraphics[height=0.43\textheight]{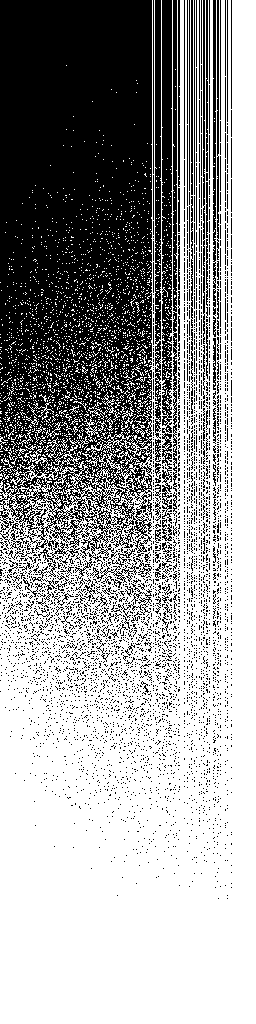}}}
            \caption{Iteration 4: 10 more}
        \end{subfigure}\hfill
        \begin{subfigure}[t]{0.15\textwidth}
            {\color{red}\fbox{\includegraphics[height=0.43\textheight]{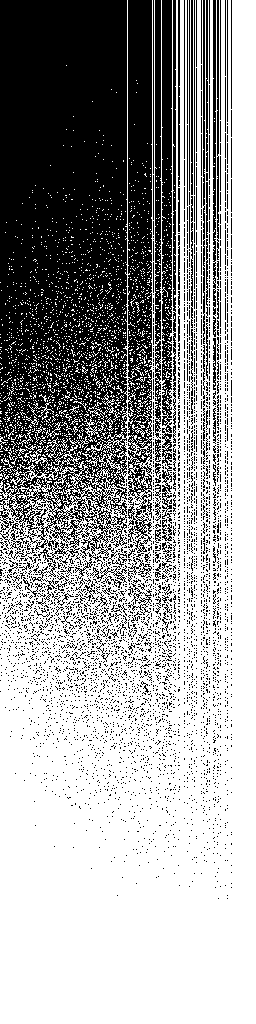}}}
            \caption{Iteration 5: 3 more; stalls}
        \end{subfigure}%
    }
    \caption{Activation mask grid of a batch ($N=1024$, $B=256$, $s=1.00$) at successive iterations of the independent trap-weight attack. A dot indicates that neuron $j$ fires on sample $i$. Rows are ordered by decreasing firing density (dense rows at the top) and columns are ordered so that the most atypical samples appear on the right. The cascade removes $17$, $18$, $13$, $10$ and $3$ samples across five iterations and then stalls, recovering $61/256 = 0.238$ of the batch—matching the recovery rate of the actual attack on this network. Peeling concentrates on the sparse bottom-right region where rare neurons intersect atypical samples.}
    \label{fig:board}
\end{figure*}

Figure~\ref{fig:board} displays such a grid recorded from a real training batch of the testbed network ($N=1024$, $B=256$, $s=1.00$) at successive iterations of the independent trap-weight attack. Rows are ordered by decreasing activation density and columns are ordered so that the most atypical samples appear on the right. Atypicality of sample $i$ is quantified by
\begin{align*}
	\operatorname{atypicality}(i)
	&= \sum_{j\in N_i}\frac{1}{p_j},\\
	p_j
	&= \frac{d_j}{B},
\end{align*}
where $N_i$ is the set of neurons activated by sample $i$ and the $p_j$ are computed once on the initial grid, before any column is deleted.
In other words, a sample is atypical when it activates neurons that themselves fire infrequently.

Replaying the peeling process on the binary grid alone—locating singleton rows, deleting the corresponding column, and repeating—removes $17$, $18$, $13$, $10$ and $3$ columns across five iterations and then stalls, recovering exactly $61$ of the $256$ samples (recovery rate $0.238$). Every column traversed by the cascade is recoverable; every column never traversed remains unrecovered.

The deletions concentrate on the sparse bottom rows of the grid and overwhelmingly on the atypical columns stacked to the right. The samples recovered in the earliest iterations are therefore the atypical members of the batch. Visually they appear darker and off-distribution relative to typical images (Figure~\ref{fig:seeds}). Thus the cascade is seeded by the atypical samples.

\section{Conclusion and Future Work}
\label{sec:conclusion}

Prior isolation-based analytic gradient-inversion attacks recover only isolated samples and are therefore bounded by the number of vertices of the convex hull of the batch~\cite{diana2025cutting}. We showed that this limitation can be overcome by an iterative peeling process inspired by Luby Transform codes: once isolated samples are recovered and subtracted from the observed gradient, previously non-isolated samples become isolated and can themselves be recovered. To maximize the cascade we further shape the first-layer activation degrees toward the robust soliton distribution via two constructions, Soliton-Free (no auxiliary data) and Soliton-Data (one auxiliary batch). Experiments on eight image and tabular data sets demonstrate that the resulting attacks recover large batches exactly, together with their labels, from a single FedSGD round—near-complete recovery ($94$--$100\%$) at batch sizes up to $128$ in the passive setting and more than $90\%$ at batch sizes of several hundred in the active setting—while certifying every recovery without ground-truth data.

Future work includes extending the cascade to FedAvg, deriving rigorous recovery guarantees under realistic data distributions, designing lightweight defenses that prevent the formation of degree-1 neurons, and investigating whether similar peeling attacks apply to networks with convolutional first layers.

\bibliographystyle{IEEEtran}
\bibliography{mybib}

\end{document}